\documentclass[sigconf]{acmart}
\AtBeginDocument{%
  }

\copyrightyear{2026}
\acmYear{2026}
\setcopyright{cc}
\setcctype{by}
\acmConference[KDD '26]{Proceedings of the 32nd ACM SIGKDD Conference on Knowledge Discovery and Data Mining V.1}{August 09--13, 2026}{Jeju Island, Republic of Korea}
\acmBooktitle{Proceedings of the 32nd ACM SIGKDD Conference on Knowledge Discovery and Data Mining V.1 (KDD '26), August 09--13, 2026, Jeju Island, Republic of Korea}
\acmPrice{}
\acmDOI{10.1145/3770854.3780283}
\acmISBN{979-8-4007-2258-5/2026/08}

\usepackage{multirow}
\usepackage{algorithm}
\usepackage{algpseudocode}
\usepackage{enumitem}
\usepackage{color}
\definecolor{my_red}{RGB}{222, 106, 81}
\definecolor{my_blue}{RGB}{146, 216, 255}
\definecolor{my_green}{RGB}{169, 209, 142}
\usepackage{subfig}
\usepackage{graphicx}
\usepackage{colortbl}
\usepackage[most]{tcolorbox}
\usepackage{balance}

\begin{document}
\newcommand{\TheName}{\textbf{HyFunc}}
\newcommand{\LML}{\text{LM\textsubscript{L}}}
\newcommand{\LMS}{\text{LM\textsubscript{S}}}

\title[\TheName{}: Accelerating LLM-based Function Calls for Agentic AI ...]{\TheName{}: Accelerating LLM-based Function Calls for Agentic AI through Hybrid-Model Cascade and Dynamic Templating}


\author{Weibin Liao}
\authornote{Work done during internship at Microsoft Corporation.}
\orcid{0000-0002-9682-9934}
\affiliation{%
  \institution{Peking University}
  \city{Beijing}
  \country{China}
}

\author{Jian-guang Lou}
\affiliation{%
  \institution{Microsoft Corporation}
  \city{Beijing}
  \country{China}
}

\author{Haoyi Xiong}
\authornote{W. Liao and H. Xiong contributed equally to this research.}
\authornote{H. Xiong is corresponding author.}
\affiliation{%
  \institution{Microsoft Corporation}
  \city{Beijing}
  \country{China}
}
\email{haoyi.xiong.fr@ieee.org}

\renewcommand{\shortauthors}{Weibin Liao, Jian-guang Lou, and Haoyi Xiong}

\begin{abstract}
While agentic AI systems rely on LLMs to translate user intent into structured function calls, this process is fraught with computational redundancy, leading to high inference latency that hinders real-time applications. This paper identifies and addresses \emph{three key redundancies}: (1) the redundant processing of a large library of function descriptions for every request; (2) the redundant use of a large, slow model to generate an entire, often predictable, token sequence; and (3) the redundant generation of fixed, boilerplate parameter syntax. We introduce \TheName{}, a novel framework that systematically eliminates these inefficiencies. \TheName\ employs a \emph{hybrid-model cascade} where a large model distills user intent into a single ``soft token.'' This token guides a lightweight retriever to select relevant functions and directs a smaller, prefix-tuned model to generate the final call, thus avoiding redundant context processing and full-sequence generation by the large model. To eliminate syntactic redundancy, our ``dynamic templating'' technique injects boilerplate parameter syntax on-the-fly within an extended vLLM engine. 
To avoid potential limitations in generalization, we evaluate \TheName\ on an unseen benchmark dataset, BFCL. Experimental results demonstrate that \TheName\ achieves an excellent balance between efficiency and performance. It achieves an inference latency of 0.828 seconds, outperforming all baseline models, and reaches a performance of 80.1\%, surpassing all models with a comparable parameter scale. These results suggest that \TheName\ offers a more efficient paradigm for agentic AI.
Our code is publicly available at \url{https://github.com/MrBlankness/HyFunc}.
\end{abstract}



\maketitle
\newcommand\kddavailabilityurl{https://doi.org/10.5281/zenodo.18137443}
\ifdefempty{\kddavailabilityurl}{}{
\begingroup\small\noindent\raggedright\textbf{Resource Availability:}\\
The source code of this paper has been made publicly available at \url{\kddavailabilityurl}.
\endgroup
}

\begin{figure}[!t]
  \centering
  \includegraphics[width=\linewidth]{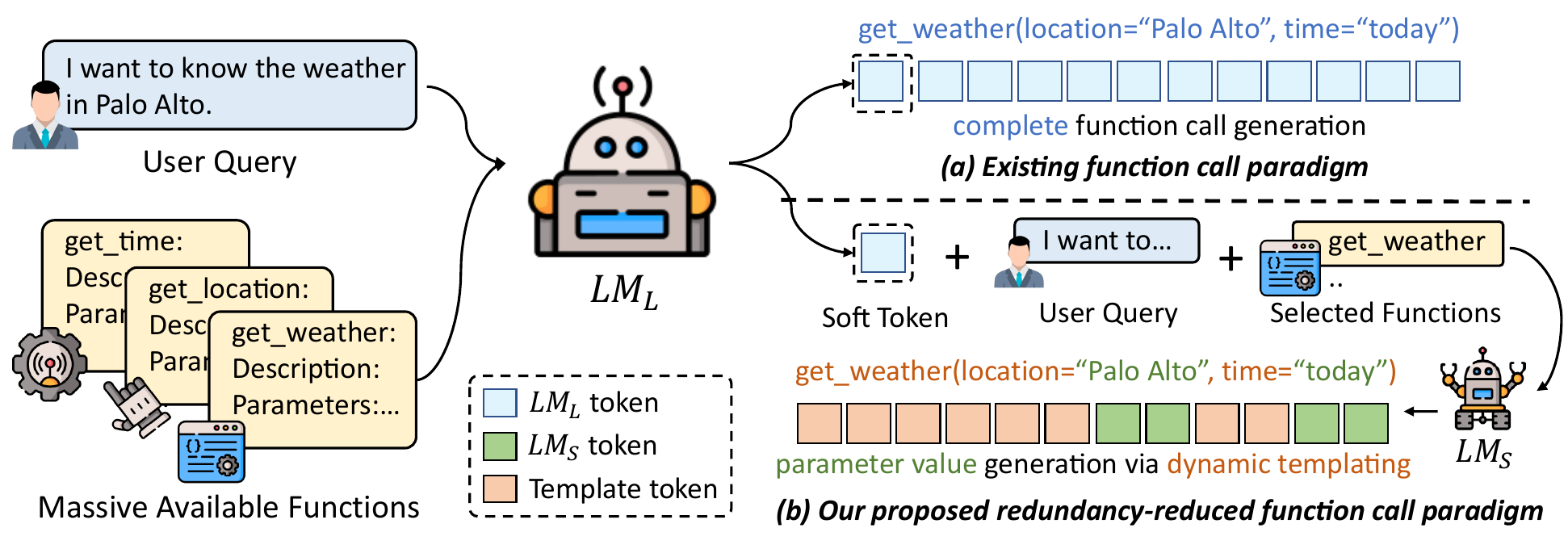}
  \caption{Illustration of the key differences between existing function call paradigm and our proposed redundancy-reduced paradigm.}
  \label{fig:teaser}
\end{figure}

\section{Introduction}\label{sec:intro}
The proliferation of agentic AI systems~\cite{acharya2025agentic,li2025towards}, capable of performing tasks by interacting with external tools~\cite{liao2025LearNAT} and APIs, marks a paradigm shift in human-computer interaction~\cite{gpt-4o,Qwen,GLM4}. At the core of these systems are LLMs that translate natural language instructions into executable function calls~\cite{BFCL,wang2025hammerbench} (e.g., \texttt{get\_weather(location="Palo Alto")}). While larger models demonstrate superior reasoning for this task, their deployment is hampered by significant inference latency~\cite{yuan2025mitigating,chen2024livemind}, creating a barrier for responsive user experiences.

We posit that a primary source of this inefficiency is \textbf{information redundancy} inherent in the standard function-calling workflow. This redundancy manifests in several ways, presenting distinct technical challenges:

\begin{enumerate}
[leftmargin=*,itemsep=0pt,parsep=0.5em,topsep=0.3em,partopsep=0.3em]
    \item \textbf{Redundancy in Context Processing:} Models are often prompted with extensive function libraries~\cite{BFCL,ToolACE,Hammer}, forcing an inefficient processing of irrelevant context on every turn. While a large model (\LML) excels at selecting the correct function, using it for the entire subsequent generation is computationally excessive. The challenge is to leverage the \LML's reasoning for selection without incurring its full generative cost.

    \item \textbf{Redundancy in Full-Sequence Generation:} Using an expensive \LML\ to generate a structured function call~\cite{liu2024apigen} is often unnecessary, as the core intent is often embedded in the first token of the generated sequence~\cite{wang2024large}. We hypothesize that the initial ``soft token''~\cite{ning2023all,kong2022spvit,liao2025PAI} from an \LML\ contains a sufficiently rich signal to guide a more efficient smaller model (\LMS). The challenge is to steer the \LMS\ effectively using only this compressed semantic signal.

    \item \textbf{Redundancy in Syntactic Generation:} A large part of a function call is fixed, boilerplate syntax (e.g., function names, parentheses, parameter names). Generating these predictable tokens is wasteful, as the core task is extracting the variable parameter \textit{values} from the context. The challenge is to bypass this redundant syntax generation and focus the model on the crucial task of value extraction.
\end{enumerate}

To address these challenges, we introduce \TheName{}, a framework designed to systematically reduce these redundancies. Our core intuition is that the function-calling process can be decomposed into two distinct phases: \emph{high-level semantic reasoning} (i.e., understanding user intent and selecting the correct function from a large function library) and \emph{low-level structured generation} (i.e., filling in parameter values when a specific function has been selected). While large models excel at the former, they are computationally excessive for the latter. Conversely, smaller models \LMS{} are efficient but may lack the reasoning power for accurate function selection. Therefore, \TheName{} is designed as a hybrid-model cascade that leverages the best efforts of both worlds. As shown in Figure.~\ref{fig:teaser} (in comparison to the existing function call paradigm), our proposal uses a large model \LML{} for a single, efficient forward pass to distill user intent as the \emph{``first soft token''}, and then hands off the task to a specialized smaller model \LMS{} for the final, constrained generation. To achieve the goal, three key strategies -- \emph{efficient function retrieval via soft token distillation}, \emph{guided generation with soft token continuous prompting}, and \emph{accelerating syntax generation with dynamic templating} have been proposed. Specifically, our key contributions are as follows.
\begin{itemize}
[leftmargin=*,itemsep=0pt,parsep=0.5em,topsep=0.3em,partopsep=0.3em]
    \item  We study the LLM function-calling acceleration problem from the perspective of redundancy in generating the tool calls from conversation context with the user, systematically identifying and targeting three key inefficiencies: the redundant processing of large function libraries, the redundant generation of the full token sequence by a large model, and the redundant creation of boilerplate syntax.

    \item We design and implement \TheName, a novel framework built on a \textbf{hybrid-model cascade} centered on the \emph{``first soft token''} generated by \LML\ to target these redundancies. It features three core components: (1) a \emph{soft token-based function retriever} that uses a single forward pass from a \LML\ to drive an efficient Multi-layer Perceptron (MLP) for function selection; (2) a smaller model (\LMS) specifically \emph{fine-tuned via prefix continuous prompting} on a dataset where the soft token from the \LML\ serves as the conditional prompt and the annotated tool call is the generation target; and (3) a \emph{dynamic templating mechanism} built into vLLM to inject boilerplate syntax on-the-fly.

    \item We conducted extensive experiments to validate the effectiveness of our approach. To avoid potential limitations in generalization, we evaluated \TheName\ on an unseen benchmark dataset, BFCL. Experimental results show that \TheName\ achieves an overall accuracy of \textbf{80.1\%}, significantly lifting the performance of a 0.6B parameter model from 62.2\% and outperforming many larger public models. This high accuracy is achieved with an end-to-end inference time of just \textbf{0.828s}, which surpasses the inference latency of all baseline LLMs.
\end{itemize}
Our framework provides a pragmatic path toward building fast, accurate, and scalable agentic systems.

\section{The \TheName\ Framework}\label{sec:method}
The \TheName\ framework is designed to accelerate function calling by decomposing the task and delegating sub-tasks to the most suitable component. This section details its architecture and the key mechanisms that drive its efficiency.

\paragraph{\textbf{Problem Formulation}}
The task of function calling is formally defined as follows. Given a user query, denoted by $q$, and a set of candidate functions $\mathbf{F} = \{f_1, f_2, \cdots, f_N\}$, where $N$ is the total number of available functions, the objective is to select the most appropriate function or sequence of functions to execute and to determine the correct arguments for each. Specifically, the model is tasked with generating a sequence of function calls:

\begin{equation}\label{equ:promble_formulation}
    \mathbf{A}=[f_1(a_1),\ldots,f_m(a_m)] = \theta_{LM}(\langle q,\mathbf{F}\rangle),
\end{equation}

where $\mathbf{A}$ represents the resulting sequence of $m$ function invocations. For the $j$-th invocation ($1 \leq j \leq m$), $f_j \in \mathbf{F}$ is the selected function and $a_j$ represents its corresponding arguments. The term $LM_{\theta}(\cdot)$ denotes the Large Language Model, parameterized by $\theta$, which processes the query $q$ and the set of function definitions $\mathbf{F}$ to produce the output.

\begin{figure}[!t]
  \centering
  \includegraphics[width=\linewidth]{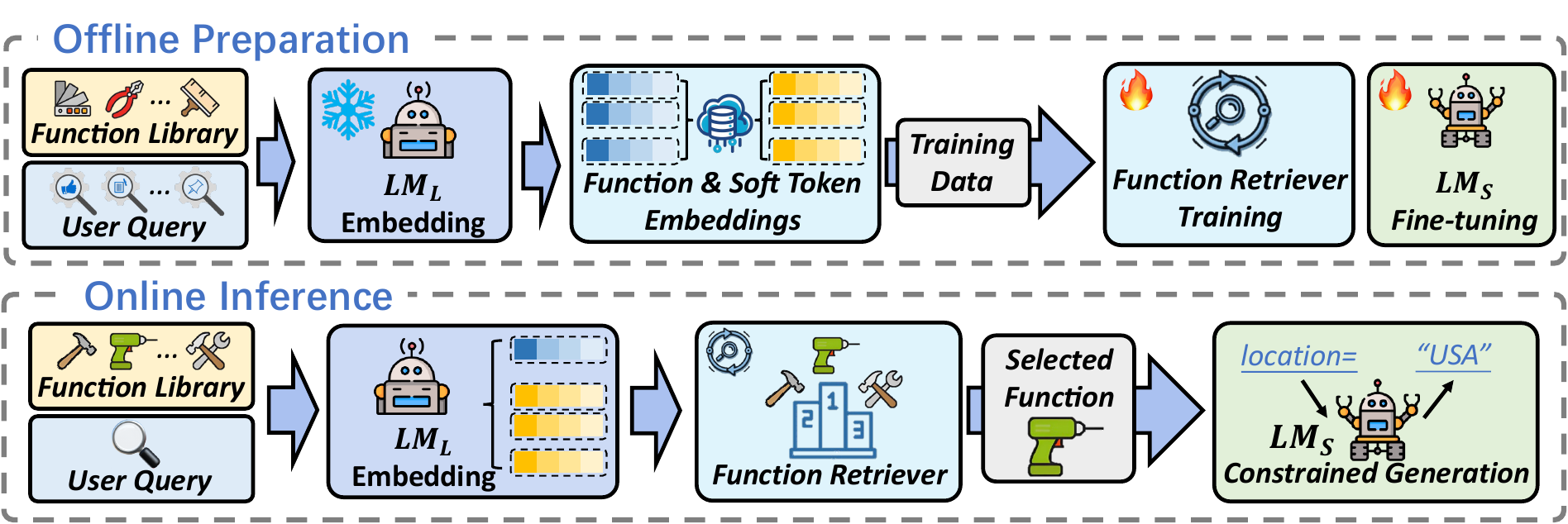}
  \caption{High-level architecture of the \TheName\ framework, including Offline Preparation and Online Inference, showing the flow from user prompt to the final function call through the two models and retrieval step.}
  \label{fig:framework}
\end{figure}

\subsection{The Hybrid-Model Cascade Architecture}
At its core, \TheName\ operates as a hybrid-model cascade that synergizes a large, powerful model (\LML) with a smaller, efficient model (\LMS). The central idea is to decouple high-level semantic reasoning from low-level structured generation. As show in Figure.~\ref{fig:framework}, the process consists of two stages:
\begin{itemize}
[leftmargin=*,itemsep=0pt,parsep=0.5em,topsep=0.3em,partopsep=0.3em]
    \item \textbf{Offline Preparation:} Before handling any user requests, we prepare the necessary components. This involves creating a semantic embedding for every function in the tool library using the \LML\ and training both the lightweight MLP-based function retriever $\mathscr{R}_\theta$ and the specialized \LMS.
    \item \textbf{Online Inference Pipeline:} When a user prompt is received, \TheName\ executes a three-phase pipeline orchestrated by the ``first soft token'' distilled from the \LML. This pipeline includes (1) efficient function retrieval, (2) guided function-call generation, and (3) accelerated syntex generation via dynamic templating. Each phase is designed to systematically eliminate one of the redundancies identified in Section.~\ref{sec:intro}.
\end{itemize}

\begin{figure*}
     \subfloat[Func. Embed. \& Soft Token Distill. via \LML\label{fig:LML_embedding}]{\includegraphics[width=0.28\linewidth]{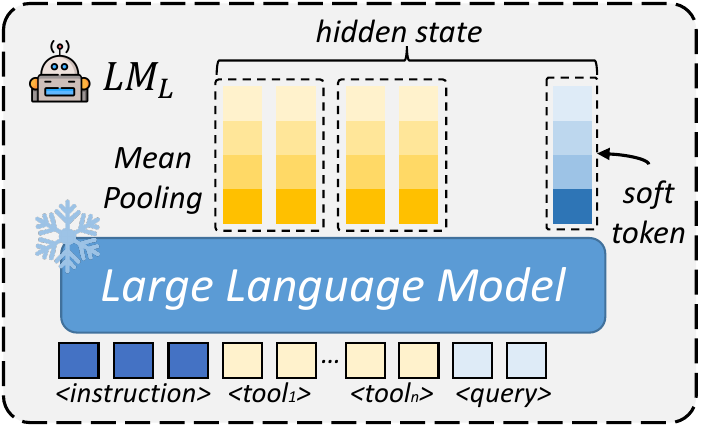}
}\
     \subfloat[Function Retriever Training \& Inference\label{fig:retriever}]{\includegraphics[width=0.35\linewidth]{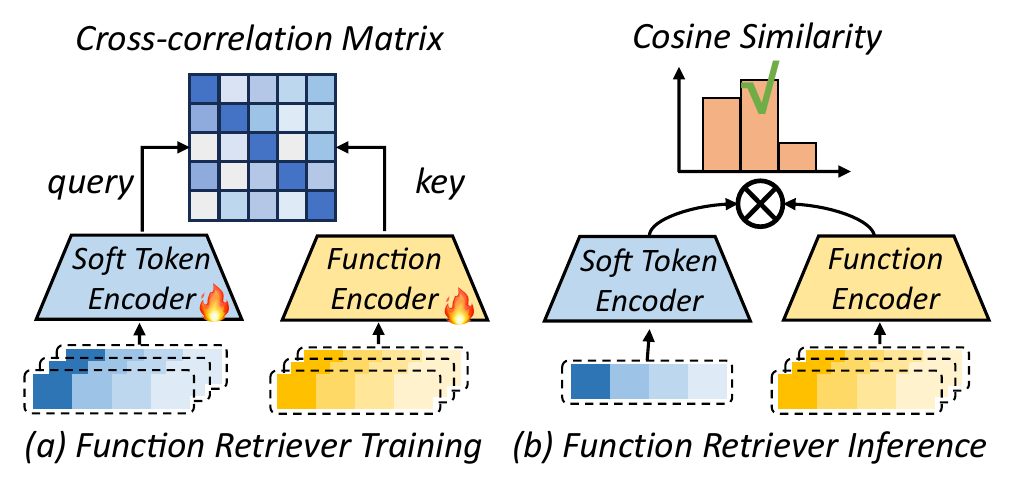}
}\
     \subfloat[Tuning \LMS{} for Guided Func. Call Gen.\label{fig:LMS_tuning}]{\includegraphics[width=0.28\linewidth]{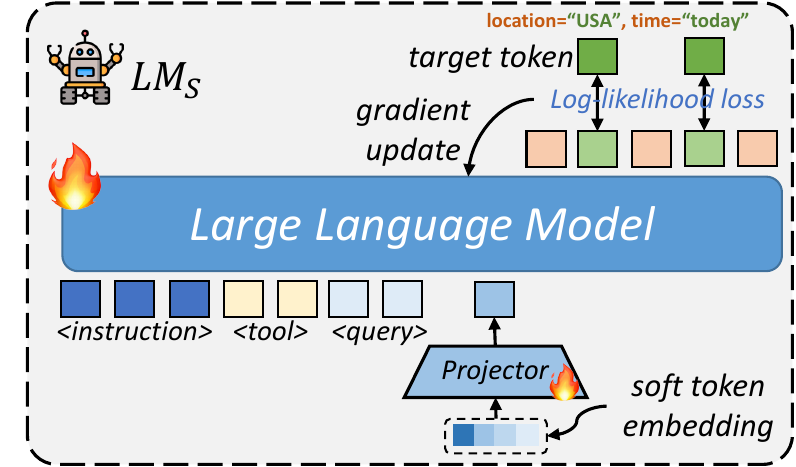}
}
    \caption{Illustration of the three core strategies in the \TheName{} framework. \textbf{(a)} The \LML\ performs a single forward pass to produce semantic embeddings: function embeddings are derived via mean pooling over their token hidden states, while the user's intent is distilled into the hidden state of the first generated ``soft token''. \textbf{(b)} A dual-encoder MLP-based retriever is trained with a contrastive loss to align soft token and function embeddings. During inference, it uses cosine similarity to efficiently select relevant functions. \textbf{(c)} The \LMS\ is fine-tuned using the soft token as a continuous prompt. A projector maps the soft token from the \LML's space to the \LMS's space, guiding the smaller model to generate the final, structured function call with reduced context.}\label{fig:three_strategies}
\end{figure*}

\subsection{Function Retrieval via Soft Token Distillation}
To address the redundancy in context processing, \TheName\ replace full-prompt reasoning with a highly efficient retrieval step. This process involves two key steps:
\begin{itemize}
[leftmargin=*,itemsep=0pt,parsep=0.5em,topsep=0.3em,partopsep=0.3em]
    \item \textbf{Offline Function \& Soft Token Embedding:} \TheName\ create a semantic signature $f_{sig}$ for every function in the tool library and encapsulates the user's intent via a single ``first soft token''.  Each function's description and user query is fed to the \LML, and \TheName\ extract the hidden states to obtain the semantic embedding $e_f$ of function and soft token embedding $e_q$ of user intent. These embeddings are stored in a fast-access database $\mathcal{D}$ as a dense embedding of the function's semantics.
    \item \textbf{Online Function Retrieval:} During inference, a single forward pass on the user's prompt through the \LML\ produces a ``first soft token'' that encapsulates the user's intent. This token serves as a query to a lightweight, pre-trained MLP retriever. The MLP rapidly compares the query token against the pre-computed function embeddings and returns a small, ranked subset of the most relevant functions, effectively pruning the search space with minimal computational overhead.
\end{itemize}

\subsubsection{\textbf{Offline Function \& Soft Token Embedding}}

\TheName\ initiates the function retrieval process by creating a semantic signature for each function within a function library, denoted as $\mathbf{F} = \{f_1, f_2, \dots, f_n\}$, and for the user's query $q$. For every function $f_i \in \mathbf{F}$, characterized by its name, description, and parameter specifications, \TheName\ constructs a detailed prompt. This is accomplished by concatenating a manually composed instruction $I$ (see Appendix.~\ref{appendix:prompt}), the comprehensive function set $\mathbf{F}$, and the specific user query $q$. 
As show in Figure.~\ref{fig:LML_embedding}, the resulting concatenated text $[I, \mathbf{F}, q]$ is tokenized as a token squeeze and subsequently input into \LML. 
From the output of \LML, \TheName\ extracts the sequence of hidden states, $H_{f_i} = \{h_1, h_2, \dots, h_k\}$, corresponding to the $k$ tokens that represent the function $f_i$. 
A \texttt{Mean Pooling} operation is then applied to these hidden states to compute the definitive function embedding, $\mathbf{e}_{f_i} \in \mathbb{R}^d$, where $d$ is the dimension of the hidden state of \LML.

Simultaneously, to distill the user's intent, \TheName\ captures the hidden state of the ``first soft token'' generated by \LML\ without decoding it. This ``soft token'' hidden state is designated as the user query embedding, $\mathbf{e}_q \in \mathbb{R}^d$. 
Upon computation, both the set of function embeddings $\{\mathbf{e}_{f_i}\}_{i=1}^N$ and the soft token query embeddings $\mathbf{e}_q$ are systematically stored in a high-speed database, $\mathcal{D}$, to facilitate the efficient training of the function retriever.

\subsubsection{\textbf{Online Function Retrieval}}
\TheName\ uses a lightweight MLP-based function retriever $\mathscr{R}_\theta$ to quickly retrieve relevant functions for a given user query.
As show in Figure.~\ref{fig:retriever}, for the training of the function retriever $\mathscr{R}_\theta$, \TheName\ implements a dual-encoder model. This architecture consists of a soft token encoder, $E_q$, and a function encoder, $E_f$, where each encoder is structured as a two-layer MLP. The training regimen utilizes the embeddings stored in the database $\mathcal{D}$. 
\TheName\ employs a contrastive learning approach~\cite{hadsell2006dimensionality,li2025rankelectra}, specifically the InfoNCE loss~\cite{oord2018representation,he2020momentum,li2025fultr}, to optimize the retriever. 
For a given query embedding $\mathbf{e}_q$, the corresponding ground-truth function embedding $\mathbf{e}_{f^+}$ serves as the positive sample, while all other function embeddings in the batch act as negative samples. The objective is to minimize the following loss function:
\begin{equation}\label{equ:infornce_loss}
\mathcal{L}_{\text{InfoNCE}} = -\log \frac{\exp(\text{sim}(E_q(\mathbf{e}_q), E_f(\mathbf{e}_{f^+})) / \tau)}{\sum_{j=1}^{B} \exp(\text{sim}(E_q(\mathbf{e}_q), E_f(\mathbf{e}_{f_j})) / \tau)},
\end{equation}
where $\text{sim}(\cdot, \cdot)$ is the cosine similarity, $\tau$ is a temperature hyperparameter, and $B$ is the number of samples in the batch.

During the inference stage, for a new query, \TheName\ first acquires the soft token embedding $\mathbf{e}_q$ and the set of pre-computed function embeddings $\{\mathbf{e}_{f_i}\}_{i=1}^N$. The trained function retriever $\mathscr{R}_\theta$ then deeply encodes these initial embeddings to yield refined representations:
\begin{equation}
\mathbf{z}_q = E_q(\mathbf{e}_q),~~~~\mathbf{z}_{f_i} = E_f(\mathbf{e}_{f_i})
\end{equation}
The relevance of each function $f_i$ is quantified by the cosine similarity between the final query and function representations:
\begin{equation}\label{equ:cosine}
s_i = \frac{\mathbf{z}_q \cdot \mathbf{z}_{f_i}}{\|\mathbf{z}_q\| \|\mathbf{z}_{f_i}\|}
\end{equation}
\TheName\ retrieves all functions $\{f_i\}$ for which the similarity score $s_i$ is greater than a specified threshold $\alpha$. Furthermore, as our subsequent experiments assume the guaranteed presence of a relevant function within the library $\mathbf{F}$ for any query, a fallback mechanism is instituted. If no function's similarity score exceeds $\alpha$, \TheName\ retrieves the single function with the highest similarity score, ensuring a candidate is always returned.

\subsection{Guided Function Call Generation with Soft Token Continuous Prompting}
With a small set of candidate functions retrieved, the task is handed to the smaller model, \LMS. To ensure the \LMS\ can effectively interpret the semantic signal from the \LML, it is specifically prepared using prefix continuous prompting~\cite{liu2022p,li2021prefix}.
\begin{itemize}
[leftmargin=*,itemsep=0pt,parsep=0.5em,topsep=0.3em,partopsep=0.3em]
    \item \textbf{Specialized Fine-Tuning:} We create a dataset where each entry consists of instruction, user query, selected functions, the corresponding soft token generated by the \LML, and the ground-truth tool call. The \LMS\ is then fine-tuned on this dataset.
    \item \textbf{Conditional Generation:} This tuning process teaches the \LMS\ to treat the high-dimensional soft token as a powerful conditional prefix that guides generation. It aligns the \LMS\ with the \LML's semantic space, enabling it to accurately generate the final tool call by conditioning on the soft token, the original prompt, and the retrieved function descriptions, thus avoiding redundant reasoning.
\end{itemize}

\subsubsection{\textbf{Specialized Fine-Tuning via Conditional Generation}}

In the second phase, \TheName\ leverages the distilled user intent, encapsulated in the soft token embedding, to guide a more efficient, smaller large language model \LMS, in generating the precise function call syntax. The core task of this phase is to perform a prefix-prompt-based fine-tuning of \LMS, where the soft token embedding from Phase 1 serves as a continuous prompt.

A significant challenge arises from the fact that the soft token embedding, $\mathbf{e}_q \in \mathbb{R}^d$, was generated by the initial model \LML, rendering it semantically misaligned with the embedding space of the smaller model \LMS~\cite{llava,minigpt}. A direct injection of $\mathbf{e}_q$ into \LMS\ would be ineffective due to this representational disparity. To bridge this semantic gap, as shown in Figure.~\ref{fig:LMS_tuning}, \TheName\ incorporates a Projector, $\mathscr{P}_\theta$, which is a single linear layer. Drawing inspiration from the LLaVA~\cite{llava} architecture, this projector maps the soft token embedding from the source space of \LML\ to the target embedding space of \LMS. Let the embedding dimension of \LMS\ be $d'$. The $\mathscr{P}_\theta$ is performed by a weight matrix $W_p \in \mathbb{R}^{d' \times d}$:
\begin{equation}\label{equ:projector}
\mathbf{p}_q = \mathscr{P}_\theta(\mathbf{e}_q) = W_p \mathbf{e}_q
\end{equation}
where $\mathbf{p}_q \in \mathbb{R}^{d'}$ is the resulting projected soft token, now acting as a continuous prompt that is comprehensible to \LMS.

For the fine-tuning process, \TheName\ constructs a new input sequence for \LMS. This sequence is formed by concatenating a manually written instruction $I$, the refined set of candidate functions $\mathbf{F}' \subseteq \mathbf{F}$ selected in Phase 1, the original user query $q$, and the projected soft token $\mathbf{p}_q$. The continuous prompt $\mathbf{p}_q$ is prepended to the token embeddings of the textual input, guiding the subsequent generation process.

The model is then optimized using a standard Supervised Fine-Tuning (SFT) objective. Given a ground-truth function call output $Y = (y_1, y_2, \dots, y_m)$, \TheName\ trains the parameters of both the smaller model \LMS\ and the Projector $\mathscr{P}_\theta$ by minimizing the negative log-likelihood of the target sequence. The loss function is defined as:
\begin{equation}\label{equ:sft_loss}
\mathcal{L}_{\text{SFT}} = - \sum_{t=1}^{m} \log P(y_t | C, y_{<t}; \theta_{LM_S}, \theta_\mathscr{P})
\end{equation}
where $C$ represents the prepared input context (including $I$, $\mathbf{F}'$, $q$, and the continuous prompt $\mathbf{p}_q$), $y_{<t}$ denotes the preceding ground-truth tokens, and $\theta_{LM_S}$ and $\theta_\mathscr{P}$ are the trainable parameters of the model \LMS\ and the projector $\mathscr{P}_\theta$, respectively. This joint optimization ensures that the projector effectively translates the guiding signal from the larger model, and the smaller model learns to interpret this signal to generate the correct function call.

\subsection{Accelerated Syntax Generation with Dynamic Templating}
The final phase of \TheName\ aims to address the redundancy in syntactic generation of function calls, targeting both the training and inference stages to maximize efficiency and accuracy. This phase is designed to reduce the redundant, token-by-token generation of predictable boilerplate syntax (e.g., function names, parentheses, and parameter keywords) and to focus the model's capabilities on the critical task of inferring parameter values.

\begin{itemize}
    [leftmargin=*,itemsep=0pt,parsep=0.5em,topsep=0.3em,partopsep=0.3em]
    \item \textbf{Selective Token Fine-Tuning:} At training time, we refine the fine-tuning process with a masked loss function. This technique, which we term Selective SFT, compels the \LMS\ to learn exclusively from the tokens that constitute parameter \textit{values}. By ignoring the static, syntactical components of a function call during backpropagation, we concentrate the model's optimization on the more challenging and crucial task of generating contextually appropriate parameter values.

    \item \textbf{Inference with Dynamic Templating:} At inference time, we deploy a novel acceleration strategy named Dynamic Templating. This method alternates between two modes: direct injection of the fixed syntactical template of a function call and controlled, auto-regressive generation by the \LMS\ for the parameter values. This process guarantees syntactically correct outputs and significantly reduces latency by bypassing LLM generation for predictable tokens.
\end{itemize}

\subsubsection{\textbf{Selective Token Fine-Tuning}}

In this phase, \TheName\ focuses on optimizing the generation of syntactically correct and semantically grounded function call parameters. A critical observation is that during the generation of a complete function call, such as \texttt{get\_weather(location="USA", time="today")}, a significant portion of the tokens constitutes a fixed template (e.g., the function name \texttt{get\_weather}, parentheses, parameter names like \texttt{location}, and assignment operators). The truly variable components that the model must infer from the user's intent are the parameter values themselves (e.g., \texttt{"USA"}, \texttt{"today"}).

To compel the \LMS\ to concentrate its learning capacity on this core inference task, \TheName\ refines the loss function used in Phase 2. Instead of calculating the loss across all tokens in the target sequence, \TheName\ employs a masked loss function that exclusively computes the error on the parameter value tokens. Let the ground-truth function call output be the token sequence $Y = (y_1, y_2, \dots, y_m)$. \TheName\ introduces a binary mask $M = (m_1, m_2, \dots, m_m)$, where $m_t=1$ if token $y_t$ is part of a parameter value, and $m_t=0$ otherwise. The new Selective SFT loss is then formulated as:
\begin{equation}
\mathcal{L}_{\text{Selective\_SFT}} = - \frac{\sum_{t=1}^{m} m_t \cdot \log P(y_t | C, y_{<t}; \theta_{LM_S}, \theta_{\mathscr{P}})}{\sum_{t=1}^{m} m_t}
\end{equation}
where $C$ is the input context. This objective effectively ignores the syntactically deterministic template components during backpropagation, thereby directing the model's optimization towards the more challenging task of generating accurate and contextually appropriate parameter values.

\subsubsection{\textbf{Inference with Dynamic Templating}}
\label{sssec:dynamic_templating}

During inference, \TheName\ employs the Dynamic Templating strategy to ensure both high-speed generation and guaranteed syntactic correctness. This process operates by dynamically switching between two distinct modes: \textbf{Template Injection Mode} and \textbf{LLM Generation Mode}.

The process begins by feeding the full context—comprising the instruction $I$, the selected function set $\mathbf{F}'$, the user query $q$, and the projected soft token $\mathbf{p}_q$—into the fine-tuned \LMS. \TheName\ then begins to construct the final function call string. In \textbf{Template Injection Mode}, \TheName\ iterates through the known structure of the selected function (its name, syntax like parentheses and commas, and parameter names). Instead of prompting the \LMS\ to generate these tokens, \TheName\ directly injects these known, correct tokens into the input sequence.

To manage the transition between modes, we introduce two special control tokens, \texttt{\textcolor{blue}{<param>}} and \texttt{\textcolor{blue}{</param>}}. The function's static template is dynamically augmented with these tokens. For instance, a function signature \texttt{get\_weather(location, time)} is transformed into the template \texttt{get\_weather(location=\textcolor{blue}{<param></param>}, \\ time=\textcolor{blue}{<param></param>})}. When the injection process encounters a \texttt{\textcolor{blue}{<param>}} token, \TheName\ switches to \textbf{LLM Generation Mode}. In this mode, control is handed to the \LMS, which, conditioned on the entire preceding context, generates the parameter value. The model has been trained via Selective SFT to conclude its value generation by emitting the \texttt{\textcolor{blue}{</param>}} token. Upon detecting this closing token, \TheName{} reverts back to \textbf{Template Injection Mode} to continue appending the rest of the fixed template. This interplay ensures that the LLM's generative power is invoked exclusively for the complex task of value inference, while the rigid syntactical structure is populated deterministically. This dual approach eliminates syntax errors and substantially reduces generation latency. 

More specifically, \TheName{} leverages the KV Cache~\cite{luohekeep} mechanism to store the representations of previously computed tokens during each mode switch. This allows the model to avoid redundant computation when generating parameter values in subsequent rounds. A detailed case study illustrating our dynamic templating strategy is provided in Section.~\ref{ssec:case_study}. The detailed algorithm is described in Appendix.~\ref{alg:dynamic_templating}.

\section{Experiments}

\subsection{Experiment Setup}

\begin{table}[!t]
\caption{
Comparison of token cost and inference time on various LLMs. 
\textbf{Input.}: \textit{Number of Input Tokens};
\textbf{Output.}: \textit{Output Token};
\textbf{Time.}: \textit{Inference Time (seconds)};
}
\label{tab:cost}
\begin{center}
\resizebox{0.9\columnwidth}{!}{
\begin{tabular}{lccc}
\toprule
\textbf{Models} & \textbf{Input} & \textbf{Output} & \textbf{Time (sec.)} \\
\midrule
\multicolumn{4}{c}{\textit{Public Models (General)}} \\
\midrule
Qwen3-8B & 621.13 & 54.68 & 1.984 \\
Qwen3-4B & 621.13 & 58.24 & 1.341 \\
Qwen3-1.7B & 621.13 & 55.26 & 1.014 \\
Qwen3-0.6B & 621.13 & 53.58 & 1.006 \\
Qwen2.5-14B-Instruct & 632.82 & 57.95 & 2.985 \\
Qwen2.5-7B-Instruct & 632.82 & 58.99 & 1.815 \\
Qwen2.5-3B-Instruct & 632.82 & 62.70 & 1.282 \\
Qwen2.5-1.5B-Instruct & 632.82 & 56.63 & 0.872 \\
Qwen2.5-0.5B-Instruct & 632.82 & 60.54 & 0.927 \\
\midrule
\multicolumn{4}{c}{\textit{Public Models (Specifical)}} \\
\midrule
ToolACE-8B & 943.05 & 55.34 & 1.876 \\
Hammer2.1-7B & 822.00 & 57.74 & 1.792 \\
Hammer2.1-3B & 822.00 & 52.98 & 1.042 \\
Hammer2.1-1.5B & 822.00 & 56.48 & 0.949 \\
Hammer2.1-0.5B & 822.00 & 52.54 & 0.904 \\
xLAM-7B-r & 810.57 & 59.65 & 1.171 \\
xLAM-1B-r & 864.03 & 65.47 & 0.911 \\
\midrule
\multicolumn{4}{c}{\TheName{}$^\clubsuit$ (\LML: ToolACE-8B, \LMS:~Qwen3-0.6B)} \\
\midrule
\textit{Total} & 1410.76 & 37.67 & 0.828 \\
\textit{ - Function Retriever} & 708.51 & 1.00 & 0.104 \\
\textit{ - Func. Call. Gen.} & 702.25 & 36.67 & 0.724 \\
\bottomrule
\end{tabular}
}
\end{center}
\end{table}
\begin{table*}[!t]
\caption{
Accuracy performance comparison on BFCL leaderboard (Out-of-Domain). 
\textbf{Sim.}: \textit{Simple}; \textbf{Mult.}: \textit{Multiple}; \textbf{Para.}: \textit{Parallel}; \textbf{P.M.}: \textit{Parallel Multiple}. 
$\dag$ and $\ddag$ indicates that our method outperforms all baseline models with $\le$ 1B and $\le$ 3B parameter, respectively.
}
\label{tab:main_results}
\begin{center}
\resizebox{0.92\textwidth}{!}{
\begin{tabular}{l|cccc|cccc|cccc|c}
\toprule
 & \multicolumn{8}{c|}{\textbf{Non-Live}} & \multicolumn{4}{c|}{\textbf{Live}} &  \\
 \cmidrule(r){2-9} \cmidrule(r){10-13}
 & \multicolumn{4}{c|}{\textbf{AST}} & \multicolumn{4}{c|}{\textbf{Execute}} & \multicolumn{4}{c|}{\textbf{AST}} &  \\
 \cmidrule(r){2-5} \cmidrule(r){6-9} \cmidrule(r){10-13}
\multirow{-3}{*}{\textbf{Models}} & \textbf{Sim.} & \textbf{Mult.} & \textbf{Para.} & \textbf{P.M.} & \textbf{Sim.} & \textbf{Mult.} & \textbf{Para.} & \textbf{P.M.} & \textbf{Sim.} & \textbf{Mult.} & \textbf{Para.} & \textbf{P.M.} & \multirow{-3}{*}{\textbf{Overall}} \\
\midrule
\multicolumn{14}{c}{\textit{Private Models}} \\
\midrule
 GPT-4o & 77.2 & 93.5 & 93.0 & 86.0 & 88.3 & 92.0 & 94.0 & 82.5 & 81.4 & 78.8 & 87.5 & 75.0 & 82.6 \\
 GPT-4-turbo & 70.4 & 91.0 & 90.0 & 87.5 & 87.4 & 90.0 & 86.0 & 77.5 & 83.7 & 78.6 & 81.2 & 70.8 & 81.0 \\
GPT-4o-mini & 74.8 & 92.0 & 90.0 & 84.0 & 83.3 & 92.0 & 84.0 & 75.0 & 78.7 & 76.2 & 87.5 & 70.8 & 79.9 \\
\midrule
\multicolumn{14}{c}{\textit{Public Models (General)}} \\
\midrule
 Qwen3-8B & 76.8 & 95.5 & 94.5 & 88.5 & 94.0 & 92.0 & 86.0 & 77.5 & 84.9 & 79.4 & 62.5 & 75.0 & 83.4 \\
 Qwen3-4B & 75.3 & 96.5 & 92.0 & 90.5 & 88.0 & 90.0 & 70.0 & 75.0 & 87.6 & 80.0 & 75.0 & 83.3 & 83.3 \\
Qwen3-1.7B & 71.1 & 93.0 & 87.5 & 81.5 & 87.0 & 88.0 & 68.0 & 75.0 & 75.6 & 72.7 & 75.0 & 75.0 & 76.9 \\
Qwen3-0.6B & 62.3 & 88.0 & 69.0 & 68.0 & 54.0 & 88.0 & 56.0 & 60.0 & 65.9 & 54.4 & 37.5 & 54.2 & 62.2 \\
 Qwen2.5-32B-Instruct & 72.8 & 94.0 & 93.5 & 88.5 & 97.6 & 88.0 & 84.0 & 77.5 & 80.2 & 80.1 & 43.8 & 62.5 & 82.2 \\
Qwen2.5-14B-Instruct & 69.7 & 95.0 & 88.0 & 89.0 & 90.4 & 92.0 & 72.0 & 85.0 & 77.1 & 75.0 & 75.0 & 70.8 & 79.0 \\
Qwen2.5-7B-Instruct & 71.8 & 95.0 & 90.0 & 86.0 & 95.4 & 94.0 & 84.0 & 77.5 & 75.6 & 75.6 & 68.8 & 66.7 & 79.6 \\
Qwen2.5-3B-Instruct & 73.3 & 92.0 & 73.5 & 76.5 & 86.9 & 90.0 & 66.0 & 70.0 & 74.0 & 72.1 & 62.5 & 45.8 & 74.9 \\
Qwen2.5-1.5B-Instruct & 72.4 & 87.0 & 81.5 & 75.5 & 88.0 & 90.0 & 78.0 & 72.5 & 74.0 & 66.1 & 50.0 & 45.8 & 72.7 \\
Qwen2.5-0.5B-Instruct & 61.2 & 78.0 & 60.0 & 50.0 & 51.2 & 88.0 & 52.0 & 52.5 & 56.2 & 41.3 & 56.2 & 20.8 & 52.4 \\
GLM4-9B-chat & 65.2 & 81.5 & 0.0 & 0.0 & 94.0 & 90.0 & 0.0 & 0.0 & 72.5 & 64.4 & 0.0 & 0.0 & 55.1 \\
\midrule
\multicolumn{14}{c}{\textit{Public Models (Specifical)}} \\
\midrule
 ToolACE-8B~\cite{ToolACE} & 76.7 & 93.5 & 90.5 & 89.5 & 97.4 & 94.0 & 88.0 & 77.5 & 73.3 & 76.7 & 81.2 & 70.8 & 81.0 \\
 Hammer2.1-7B~\cite{Hammer} & 78.1 & 95.0 & 93.5 & 88.0 & 86.4 & 92.0 & 86.0 & 77.5 & 76.7 & 77.4 & 81.2 & 70.8 & 81.6 \\
Hammer2.1-3B~\cite{Hammer} & 81.4 & 95.0 & 89.5 & 81.5 & 82.9 & 92.0 & 84.0 & 77.5 & 73.3 & 73.3 & 62.5 & 66.7 & 79.0 \\
Hammer2.1-1.5B~\cite{Hammer} & 74.7 & 92.0 & 84.5 & 80.0 & 86.6 & 90.0 & 82.0 & 75.0 & 71.3 & 69.8 & 50.0 & 62.5 & 75.5 \\
Hammer2.1-0.5B~\cite{Hammer} & 68.0 & 83.0 & 71.5 & 54.0 & 68.4 & 84.0 & 82.0 & 47.5 & 60.1 & 58.0 & 50.0 & 45.8 & 63.5 \\
 BigAgent-8B~\cite{bitagent} & 76.2 & 95.0 & 94.0 & 82.5 & 98.6 & 94.0 & 88.0 & 77.5 & 77.9 & 77.4 & 87.5 & 70.8 & 81.6 \\
xLAM-8x7B-r~\cite{xlam} & 73.6 & 90.0 & 69.0 & 38.0 & 89.0 & 90.0 & 72.0 & 45.0 & 74.8 & 79.3 & 43.8 & 58.3 & 74.3 \\
xLAM-7B-r~\cite{xlam} & 74.2 & 95.5 & 81.0 & 73.5 & 74.0 & 96.0 & 82.0 & 67.5 & 72.1 & 74.9 & 50.0 & 62.5 & 76.6 \\
xLAM-1B-r~\cite{xlam} & 71.7 & 86.0 & 5.0 & 2.0 & 77.8 & 90.0 & 4.0 & 0.0 & 64.0 & 53.4 & 6.2 & 0.0 & 51.2 \\
Granite-20B~\cite{Granite} & 72.8 & 91.5 & 84.0 & 81.5 & 84.9 & 92.0 & 86.0 & 82.5 & 68.2 & 56.3 & 43.8 & 58.3 & 78.6 \\
\midrule
\multicolumn{14}{c}{Our Proposal -- \TheName{}$^\heartsuit$ (\LML: ToolACE-8B,  \LMS:~Qwen2.5-0.5B-Instruct) and \TheName{}$^\clubsuit$ (\LML: ToolACE-8B, \LMS:~Qwen3-0.6B)} \\
\midrule
\rowcolor[RGB]{237,237,237}
\TheName $^{\heartsuit}$ & 72.5$^\dag$ & 92.5$^\dag$ & 87.5$^\dag$ & 85.0$^\dag$ & 91.0$^\dag$ & 90.0$^\dag$ & 78.0 & 75.0$^\dag$ & 74.8$^\dag$ & 75.2$^\dag$ & 68.8$^\dag$ & 66.7$^\dag$ & 78.6$^\dag$ \\
\rowcolor[RGB]{237,237,237}
\TheName $^{\clubsuit}$ & 75.3$^\ddag$ & 93.5$^\dag$ & 89.5$^\dag$ & 89.0$^\ddag$ & 95.0$^\dag$ & 92.0$^\dag$ & 82.0$^\dag$ & 75.0$^\dag$ & 75.2$^\dag$ & 75.5$^\dag$ & 75.0$^\ddag$ & 70.8$^\dag$ & 80.1$^\ddag$\\
\bottomrule
\end{tabular}
}
\end{center}
\end{table*}

\subsubsection{\textbf{Datasets, Evaluation and Models.}}

We employ the dataset\footnote{\url{https://huggingface.co/datasets/Salesforce/xlam-function-calling-60k}} provided by xLAM~\cite{xlam} as our offline preparation, which comprises 11,300 user queries. This dataset includes various types of function calling scenarios, such as those involving multiple candidate tools and parallel function executions.
We use the Berkeley Function Call Leaderboard (BFCL) dataset~\cite{BFCL} (Out-of-Domain) to evaluate the performance of \TheName\ in the online inference setting.
For BFCL, we assess both the Non-Live and Live subsets, corresponding to synthetic test cases and real-world scenarios, respectively. Each subset is further divided into four categories: Simple, Multiple, Parallel, and Parallel Multiple. The Simple and Multiple categories both involve a single invoked tool, with Multiple featuring multiple candidate tools. In contrast, Parallel and Parallel Multiple require invoking multiple tools simultaneously.
Additionally, the Non-Live subset of BFCL contains two disjoint splits, each evaluated using different metrics: AST substring matching and execution-response matching.
For more details on the evaluation protocol, we refer readers to the original BFCL dataset documentation\footnote{\url{https://gorilla.cs.berkeley.edu/leaderboard.html}}.

We utilize ToolACE-8B\footnote{\url{https://huggingface.co/Team-ACE/ToolACE-8B}} as \LML, and adopt Qwen3-0.6B~\cite{Qwen3} and Qwen2.5-0.5B-Instruct~\cite{Qwen} as \LMS. As previously described, we implement a dual-encoder architecture for our function retriever, where each encoder is a two-layer multilayer perceptron (MLP).
For \LMS\ tuning with soft tokens, we incorporate a single-layer linear projector to map the learned representations.

\subsubsection{\textbf{Implementation Details.}}
\TheName\ was implemented using the PyTorch framework with the HuggingFace Transformers~\citep{wolf2019huggingface} library. All experiments were conducted on a server equipped with 8 $\times$ NVIDIA RTX 4090 GPUs (24G). 
The training process consisted of two distinct stages with tailored hyperparameters. For the function retriever $\mathscr{R}_\theta$ training, we employed the AdamW optimizer with a learning rate of 1e-3 and a batch size of 256. This training was performed over 100 epochs, utilizing a cosine learning rate scheduler with a warm-up phase.
For the subsequent \LMS\ fine-tuning stage, we again used the AdamW optimizer, but with a learning rate of 2e-5 and a batch size of 1. This stage was conducted for a single epoch.
Across all experiments, we set the selection threshold for the function retriever to $\alpha=0.5$ and configured the temperature parameter in our contrastive loss function (see Equ.~\ref{equ:infornce_loss}) to $\tau=0.07$.

\begin{table*}[!t]
\caption{
Ablation Study of our proposed modules in \TheName. \textbf{Sim.}: \textit{Simple}; \textbf{Mult.}: \textit{Multiple}; \textbf{Para.}: \textit{Parallel}; \textbf{P.M.}: \textit{Parallel Multiple}. 
}
\label{tab:ablation_study}
\begin{center}\scriptsize
\resizebox{0.95\textwidth}{!}{
\begin{tabular}{l|cccc|cccc|cccc|c}
\toprule
 & \multicolumn{8}{c|}{\textbf{Non-Live}} & \multicolumn{4}{c|}{\textbf{Live}} &  \\
 \cmidrule(r){2-9} \cmidrule(r){10-13}
 & \multicolumn{4}{c|}{\textbf{AST}} & \multicolumn{4}{c|}{\textbf{Execute}} & \multicolumn{4}{c|}{\textbf{AST}} &  \\
 \cmidrule(r){2-5} \cmidrule(r){6-9} \cmidrule(r){10-13}
\multirow{-3}{*}{\textbf{Methods}} & \textbf{Sim.} & \textbf{Mult.} & \textbf{Para.} & \textbf{P.M.} & \textbf{Sim.} & \textbf{Mult.} & \textbf{Para.} & \textbf{P.M.} & \textbf{Sim.} & \textbf{Mult.} & \textbf{Para.} & \textbf{P.M.} & \multirow{-3}{*}{\textbf{Overall}} \\
\midrule
Qwen3-0.6B (base) & 62.3 & 88.0 & 69.0 & 68.0 & 54.0 & 88.0 & 56.0 & 60.0 & 65.9 & 54.4 & 37.5 & 54.2 & 62.2 \\
\multicolumn{1}{r|}{w/ \textit{Function Retriever}} & 62.3 & 89.0 & 69.0 & 72.5 & 54.0 & 88.0 & 56.0 & 65.0 & 65.9 & 59.2 & 37.5 & 54.2 & 64.6 \\
\multicolumn{1}{r|}{w/ \textit{Soft Token Distillation}} & 68.8 & 90.0 & 75.0 & 78.0 & 79.0 & 88.0 & 64.0 & 70.0 & 68.6 & 66.6 & 50.0 & 62.5 & 71.2 \\
\multicolumn{1}{r|}{w/ \textit{Selective Token Tuning}} & 70.5 & 90.5 & 80.5 & 82.5 & 84.0 & 90.0 & 68.0 & 72.5 & 70.5 & 68.9 & 56.3 & 66.7 & 73.9 \\
\multicolumn{1}{r|}{w/ \textit{ Dynamic Templating}} & 75.3 & 93.5 & 89.5 & 89.0 & 95.0 & 92.0 & 82.0 & 75.0 & 75.2 & 75.5 & 75.0 & 70.8 & 80.1 \\
\bottomrule
\end{tabular}
}
\end{center}
\end{table*}

\subsubsection{\textbf{Baselines.}}

We introduce three categories of baseline models: private models, general public models, and function call–specific public models.
The private modes include the GPT-4 series~\cite{gpt-4o},
the general public models include the Qwen3 series~\cite{Qwen3}, the Qwen2.5 series~\cite{Qwen}, as well as GLM4-9B-Chat~\cite{GLM4}, and the function call–specific public models include the ToolACE~\cite{ToolACE} (ICLR'25), Hammer series~\cite{Hammer} (ICLR'25), BigAgent~\cite{bitagent}, xLAM series~\cite{xlam} (NAACL'25) and Granite~\cite{Granite} (EMNLP'24), which are detailed in Appendix.~\ref{appendix:baselines}.

\subsection{Main Results}
We evaluate \TheName\ and the introduced baselines on the BFCL benchmark, \textbf{focusing on the trade-off between accuracy and inference efficiency}. The experimental results for accuracy and latency are summarized in Table~\ref{tab:main_results} and Table~\ref{tab:cost}, for time cost and accuracy, respectively.

\subsubsection{\textbf{Accuracy and Efficiency Trade-off.}}
Our primary contribution is not solely to achieve state-of-the-art accuracy, but to provide a highly efficient solution that maintains strong performance. \TheName$^\clubsuit$ achieves an inference latency of just \textbf{0.828 seconds} per case, outperforming all baseline models in terms of speed. This efficiency stems from our redundancy-reduction design, which significantly decreases the number of generated \emph{output tokens} by 29.54\% compared to its base model, Qwen3-0.6B. Although \TheName\ introduces some additional input tokens due to its hybrid-model architecture--particularly from the \textit{Function Retriever} and \textit{Function Call Generator}--this slight increase in input length does not compromise its overall inference efficiency.

When comparing models with similar time costs (i.e., latency around 1 second), \TheName$^\clubsuit$ demonstrates superior accuracy. For example, it achieves an overall accuracy of \textbf{80.1\%}, significantly outperforming other fast models such as Hammer2.1-1.5B (75.5\%), Qwen3-1.7B (76.9\%), and xLAM-1B-r (51.2\%). This observation highlights \TheName's ability to deliver high accuracy without a latency penalty. Conversely, while some much larger models achieve slightly better accuracy, they do so at a substantially higher computational cost. For instance, GPT-4o achieves 82.6\% accuracy but is a proprietary model (where the raw inference latency on same compute setting is  not accessible). Similarly, Qwen3-8B reaches 83.4\% but takes 1.984s to run--more than twice as long as \TheName. This observation demonstrates that \TheName\ occupies a unique position on the performance-efficiency frontier, making it a more practical choice for real-time applications.

\subsubsection{\textbf{Significant Performance Improvement for Small Models.}}
When integrated with the \TheName\ framework, small language models show substantial performance gains. Qwen2.5-0.5B-Instruct improves its overall accuracy to 78.6\% (\textcolor{my_red}{26.2\%$\uparrow$}), while Qwen3-0.6B reaches 80.1\% (\textcolor{my_red}{17.9\%$\uparrow$}). These results highlight the strong efficacy of \TheName\ in enhancing the capabilities of small, efficient models to the level of much larger ones.

\subsection{Ablation Study}

To analyze the contribution of each individual component in \TheName, we incrementally add the proposed modules on top of a base \LMS. We select Qwen3-0.6B as the base \LMS\ for this ablation study. The results, presented in Table.~\ref{tab:ablation_study}, show that performance steadily improves as more \TheName\ modules are introduced.
\textbf{(1) Effectiveness of Function Retrieval.}
Although our function retriever is designed to help the model recall the most relevant functions, its effectiveness is most evident in scenarios involving multiple candidate functions. For example, in the \textit{Simple} and \textit{Parallel} subsets of BFCL—where the correct function is directly provided—retrieval is unnecessary, and thus no performance gain is observed. However, in more challenging subsets requiring function selection, such as \textit{Multiple} and \textit{Parallel-Multiple}, the function retriever improves performance by 2.4\%. Furthermore, as it relies solely on two lightweight MLPs, the retriever is significantly faster and more efficient than traditional LLM-based retrieval methods.
\textbf{(2) Effectiveness of Soft Token Distillation.}
\TheName\ distills latent user intent from \LML\ to enhance the \LMS's understanding of task requirements. This module yields a substantial performance improvement of 6.6\%, demonstrating the value of distillation in aligning user intent with token-level supervision.
\textbf{(3) Effectiveness of Selective Token Tuning.}
To avoid redundant fine-tuning on fixed function syntax, \TheName\ focuses parameter optimization on key tokens, particularly those representing argument values. This selective tuning allows the model to concentrate on generating correct parameter content, resulting in a 2.7\% performance improvement.
\textbf{(4) Effectiveness of Dynamic Templating.}
\TheName\ dynamically injects function templates during inference to prevent the model from generating incorrect arguments or forgetting required parameters—issues that are especially pronounced in smaller models. By offloading template memorization, the model can better focus on generating accurate parameter values, leading to a 6.2\% improvement in performance.

\begin{table}[!t]
\caption{
Analysis of How \TheName{} Mitigates Redundancy in the Function Calling Process.
}
\label{tab:redundancy_analysis}
\begin{center}
\resizebox{\columnwidth}{!}{
\begin{tabular}{l|ccc}
\toprule
\textbf{Redundant Stage} & \textbf{Redundant token} & \textbf{\TheName{}} & \textbf{Speed Up} \\
\midrule
Context Processing & 584.46 & function embedding \& retrieve & 5.20\% \\
Full-Sequence Generation & 53.58 & LM\_L $\rightarrow$ LM\_S & 38.51\% \\
Syntactic Generation & 16.91 & LM\_L (output) $\rightarrow$ Dynamic Template (input) & 14.56\% \\
\bottomrule
\end{tabular}
}
\end{center}
\end{table}

\subsection{Redundancy Elimination Analysis}

As shown in Table.~\ref{tab:redundancy_analysis}, we analyze the redundancies arising during the function calling process, which can be categorized into three stages: Context Processing, Full-Sequence Generation, and Syntactic Generation (see Sec.~\ref{sec:intro}). Using the BFCL dataset, we quantify the average number of redundant tokens generated per function call at each stage. We further present the targeted strategies introduced by \TheName{} to eliminate these redundancies, along with the resulting relative speedups after incorporating these strategies. Experimental results demonstrate that \TheName{} effectively mitigates redundancies in the function calling process and substantially accelerates function invocation.


\begin{table}[!t]
\caption{
Performance of \textit{Function Retriever}.
\textbf{EMAcc.}: \textit{Exact Match Accuracy}; \textbf{Pre.}: \textit{Precision}; \textbf{Rec.}: \textit{Recall}; \textbf{P.P.}: \textit{Parallel Prediction}.
}
\label{tab:retriever_result}
\begin{center}
\resizebox{0.85\columnwidth}{!}{
\begin{tabular}{ll|cccc}
\toprule
\multicolumn{2}{l|}{\textbf{Dataset Category}} & \textbf{EMAcc.} & \textbf{Pre.} & \textbf{Rec.} & \textbf{F1} \\
\midrule
\multirow{2}{*}{Non-Live} & AST & 94.0 & 95.3 & 96.5 & 95.9 \\
 & Execute & 95.8 & 96.6 & 97.4 & 97.0  \\
\midrule
Live & AST & 85.8 & 87.4 & 89.6 & 88.5  \\
\bottomrule
\end{tabular}
}
\end{center}
\end{table}

\subsection{Performance of Function Retriever}

We further conduct a standalone evaluation of the \textit{Function Retriever} component, with the results presented in Table.~\ref{tab:retriever_result}. In this experiment, we employ the Exact Match Accuracy metric: Accuracy = 1 if and only if the retriever successfully recalls all required functions without predicting any irrelevant ones. (detailed definition of Exact Match Accuracy provided in Appendix.~\ref{appendix:em_acc})
Our results demonstrate that the \textit{Function Retriever} achieves high accuracy, reaching up to 95.8\% in the Non-Live (Execute) setting. This strong performance highlights the effectiveness of our retrieval mechanism, which distills user intent through soft token supervision and enables a decoder-free function retrieval strategy.
We further analyze the impact of the function retriever’s hyperparameter $\alpha$ on the performance of \TheName{} in Fig.~\ref{fig:alpha}.

\begin{figure}[!t]
 \centering
 \includegraphics[width=0.95\linewidth]{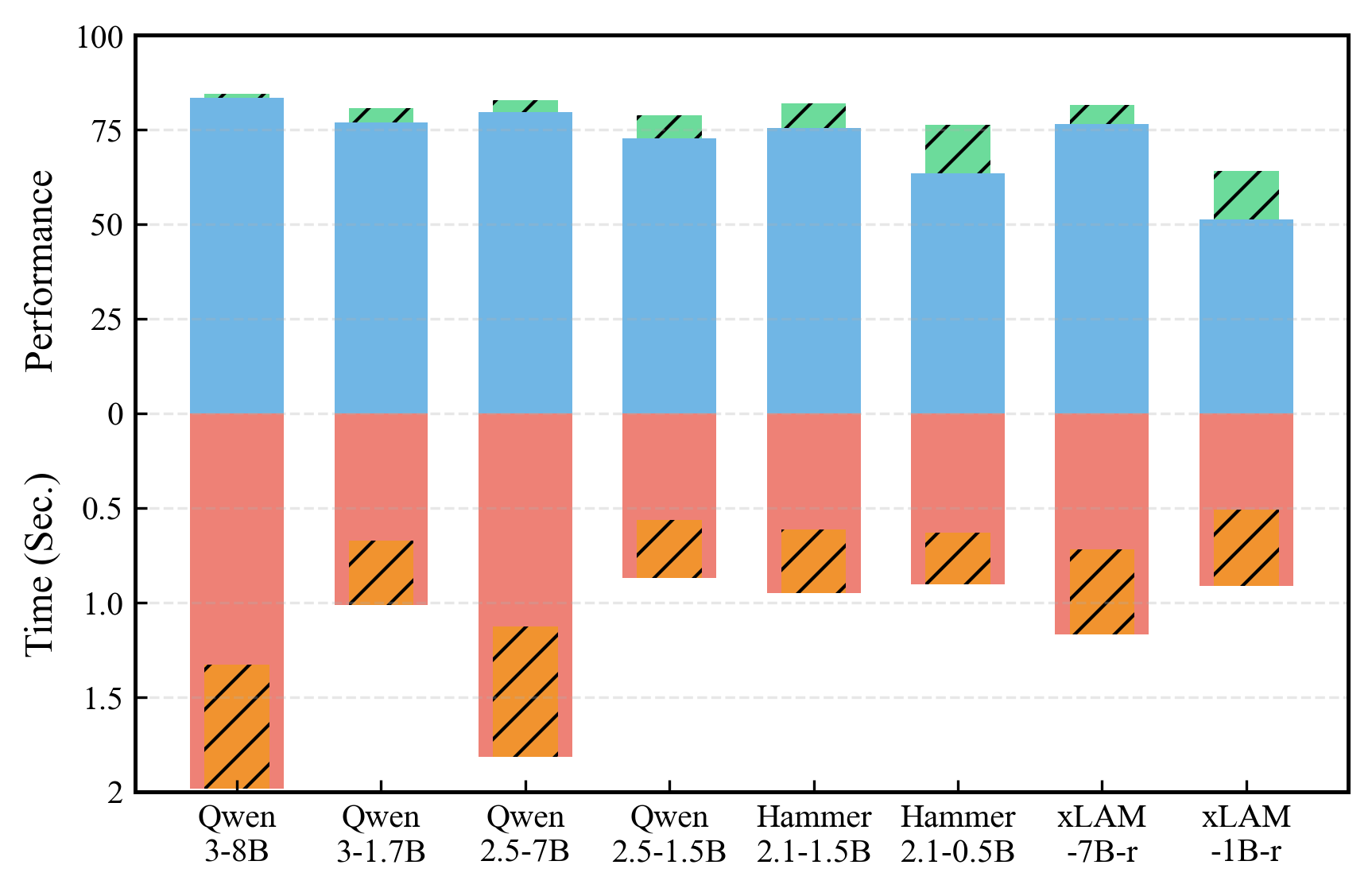}
 \caption{Performance \textit{improvement} and time consumption \textit{reduction} of \textit{Dynamic Templating} on various backbone LLMs.}
 \label{fig:dynamic_template}
\end{figure}

\subsection{Plug-and-Play of Dynamic Templating}

We propose that our \textit{Dynamic Templating} module is designed to be plug-and-play—it can be seamlessly integrated into existing function-calling models with minimal overhead. To validate this claim, we incorporate \textit{Dynamic Templating} into several baseline LLMs and evaluate its impact on function-calling performance. The results, presented in Fig.~\ref{fig:dynamic_template}, demonstrate that \textit{Dynamic Templating} consistently enhances function-calling capabilities across a wide range of models, including both general-purpose LLMs and task-specific models.
For instance, \textit{Dynamic Templating} helps Qwen2.5-1.5B-Instruct achieve an overall performance of 78.8\% (\textcolor{my_red}{6.1\%$\uparrow$}), and boosts Hammer2.1-0.5B to 76.3\% (\textcolor{my_red}{12.9\%$\uparrow$}). These results provide strong evidence of \textit{Dynamic Templating}’s effectiveness and ease of integration.
In summary, we emphasize that \textit{Dynamic Templating} plays a critical role in enhancing the reliability and usability of small-scale LLMs, making it a valuable component for real-world deployment under constrained resource settings.

\subsection{Case Study}\label{ssec:case_study}

\begin{figure}[!t]
 \centering
 \includegraphics[width=0.95\linewidth]{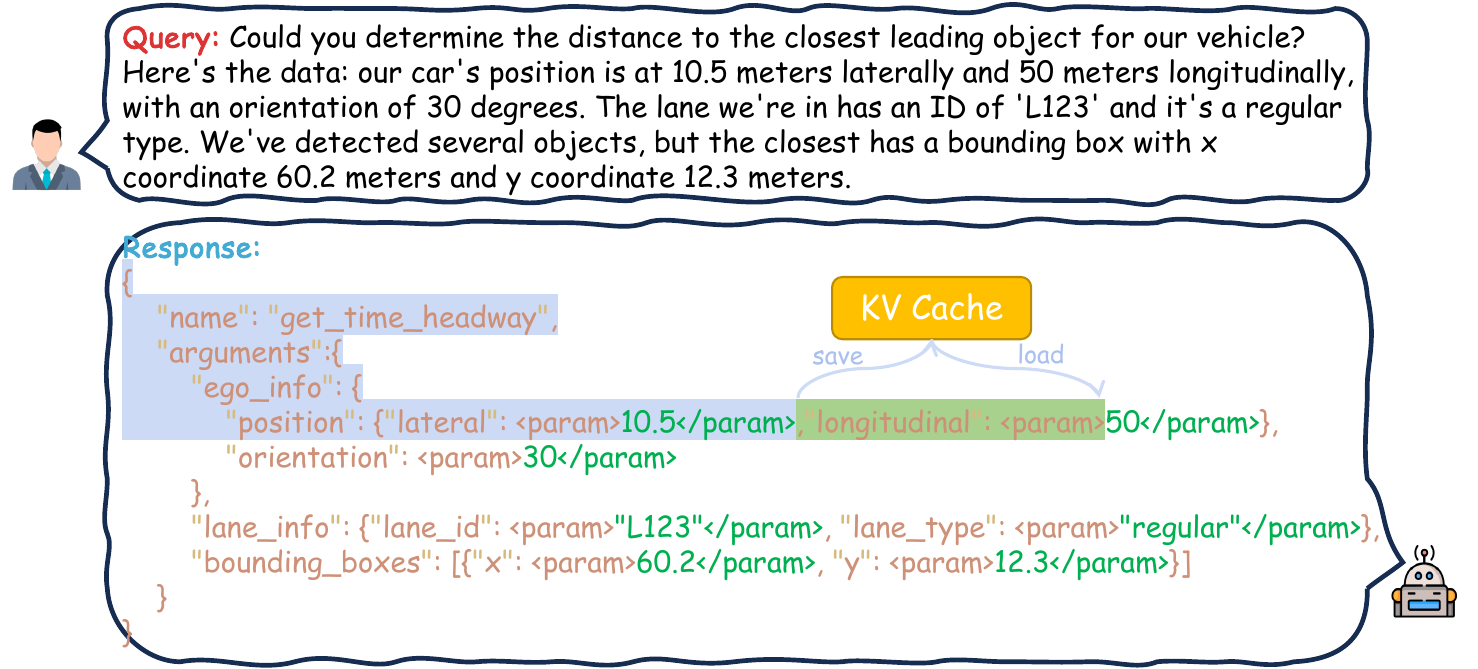}
 \caption{Case study of \textit{Dynamic Templating}.}
 \label{fig:case_study}
\end{figure}

Figure.~\ref{fig:case_study} provides a detailed illustration of the dynamic templating mechanism. The {\color[rgb]{0.815,0.608,0.510} dark brown } text denotes the function template (i.e., the input), while the {\color[rgb]{0,0.690,0.314} green } text represents \TheName’s output. \TheName\ switches between \textbf{LLM Generation Mode} and \textbf{Template Injection Mode} by monitoring the \texttt{<param>} and \texttt{</param>} tokens. During each mode switch, it leverages the KV Cache to store the representations of already processed tokens, thereby avoiding redundant attention computations. This mechanism ensures that the LLM does not generate incorrect function templates, such as syntactic errors or missing parameters. Moreover, by offloading template generation and requiring the LLM to generate only parameter values, \TheName\ significantly accelerates function invocation.

\section{Related Works}
Recent advancements have significantly expanded the function calling capabilities of large language models (LLMs), enabling them to interact with external APIs and structured data sources to perform complex, multi-step reasoning tasks~\cite{qintoolllm,liao2025TPO}. Existing approaches can broadly be categorized into tuning-free and tuning-based paradigms. Tuning-free methods leverage in-context learning through prompting strategies, such as few-shot demonstrations of API usage, without requiring additional model updates~\cite{mialon2023augmented,hsieh2023tool,ruan2023tptu}. Among these, ReAct~\cite{yao2023react} stands out for its ability to interleave reasoning and function execution in iterative cycles. In contrast, tuning-based methods aim to directly improve function invocation capabilities through task-specific fine-tuning~\cite{xlam,schick2023toolformer,ToolACE,patil2024gorilla,Granite,tang2023toolalpaca,liao2025magical}. 

To accelerate LLM inference, multi-faceted approaches have been proposed from the perspectives of models, system-level optimization, and attention mechanisms. Model-centric techniques such as quantization~\cite{lin2024awq}, network pruning, knowledge distillation to compressed models have been well surveyed in~\cite{girija2025optimizing}. These efforts are complemented by system-level optimizations that tackle core bottlenecks without altering the model itself~\cite{zheng2024sglang,triton,vllm}. For instance, FlashAttention is a pivotal I/O-aware algorithm that re-engineers the attention mechanism to minimize data movement on GPUs~\cite{dao2022flashattention}. Prefill-decode disaggregation improves LLM serving performance, enabling higher request rates and stricter latency adherence
by minimizing communication overhead~\cite{agrawal2023sarathi,zhong2024distserve,kamath2025pod,li2025rankexpert}. Architectural variants such as Grouped-Query Attention (GQA)~\cite{ainslie2023gqa}, Natively Sparse Attention (NSA)~\cite{yuan2025native}, and various KV cache strategies~\cite{li2024survey}, shrink the massive memory footprint required for long contexts. Furthermore, methods such as speculative decoding use a smaller draft model to generate multiple tokens at once, reducing latency~\cite{leviathan2023fast}. 

\paragraph{\textbf{Discussion.}} While the aforementioned works in agentic function calling have advanced the correctness of LLMs, they have largely overlooked the redundant computations inherent in the process. \emph{In contrast, our approach explicitly addresses this limitation, thereby enabling more efficient and streamlined function execution.} Compared to the broader field of LLM acceleration, our work is specifically tailored to the function-calling scenario. \emph{Our core contribution is the hybrid-model cascade architecture, which, together with our three strategies, leverages \LML-distilled soft tokens, a continuous prompt-tuned \LMS{}, and a reduced context to enhance generation.} From the perspective of accelerating function calls, our contribution is unique and effective. It is important to note that our work is orthogonal to existing efforts; it can be complemented by incorporating more advanced agentic models as either the \LML{} or \LMS, or by serving our cascade architecture with state-of-the-art inference engines to enjoy and enhance the latest research results from the both perspectives.

\section{Limitations and Future Works}

Despite the promising results obtained in our work, it is important to acknowledge the limitations.
The first limitation of \TheName{} is its sensitivity to the performance of the function retriever. As illustrated in Fig.~\ref{fig:alpha}, we vary the retriever’s hyperparameter $\alpha$ to control retrieval quality, and the results show that fluctuations in retriever performance can lead to substantial performance variance in \TheName{}. This is because retrieving incorrect or redundant functions inevitably results in erroneous function calls in subsequent stages of \TheName{}.
Moreover, the dynamic templates adopted by \TheName{} \textbf{force} the LLM to generate function arguments, which may not only produce semantically meaningless parameter values but also introduce additional redundant decoding. This issue is particularly critical in scenarios where function call is unnecessary. In contrast, the BFCL dataset mitigates this challenge, as all function-calling tasks in BFCL require at least one function calling, thereby reducing the overall difficulty faced by \TheName{}\footnote{It is worth noting that incorrect function calling is a common issue across existing function-calling approaches and is not unique to \TheName{}. By decomposing function calling into staged parsing, retrieval, and generation, \TheName{} makes such errors more explicit and transparent.}.

Another limitation is that the current implementation of \TheName{} focuses exclusively on single-turn function-calling scenarios. Some existing works, such as UserBench~\cite{qian2025userbench}, incorporate active questioning strategies to better resolve ambiguities in user intent and thus better support multi-turn interactions. In future work, we plan to further extend \TheName{} to multi-turn dialogue settings.

\section{Conclusions}
In this work, we addressed the critical challenge of inference latency in LLM-based function calling by targeting three key sources of computational redundancy. We introduced \TheName, a novel framework that employs a hybrid-model cascade to leverage the reasoning of a large model and the efficiency of a small model. Through soft token distillation, an efficient MLP-based retriever, and a dynamic templating mechanism, \TheName{} successfully reduces redundant context processing, full-sequence generation, and syntax creation. 
To validate the effectiveness of our approach and avoid potential limitations in generalization, we evaluated our method on an unseen benchmark dataset, BFCL. 
Extensive experiments demonstrate that \TheName{} achieves an excellent balance between efficiency and performance, reducing latency to just 0.828s while enabling a 0.6B parameter model to achieve 80.1\% accuracy, outperforming many larger models. This work presents a pragmatic and effective path toward deploying fast, reliable, and scalable agentic AI systems in real-world applications.



\bibliographystyle{ACM-Reference-Format}
\bibliography{references}

@string{acl = "Association for Computational Linguistics (ACL)"}

@string{emnlp = "Empirical Methods in Natural Language Processing (EMNLP)"}

@string{naacl = "North American Chapter of the Association for Computational Linguistics (NAACL)"}

@string{acl_ijcnlp = "Association for Computational Linguistics and International Joint Conference on Natural Language Processing (ACL-IJCNLP)"}

@string{colm = "Conference on Language Modeling (COLM)"}

@string{acl_findings = "Findings of Association for Computational Linguistics (ACL)"}

@string{neurips = "Advances in Neural Information Processing Systems (NeurIPS)"}

@string{icml = "International Conference on Machine Learning (ICML)"}

@string{iclr = "International Conference on Learning Representations (ICLR)"}

@string{cvpr = "IEEE Conference on Computer Vision and Pattern Recognition (CVPR)"}

@string{iccv = "International Conference on Computer Vision (ICCV)"}

@string{eccv = "European Conferenceon Computer Vision (ECCV)"}

@string{kdd = "Conference on Knowledge Discovery and Data Mining (KDD)"}

@string{sigir = "Conference on Information Retrieval (SIGIR)"}

@string{tmlr = "Transactions on Machine Learning Research (TMLR)"}

@inproceedings{li2025fultr,
  title={FULTR: A Large-Scale Fusion Learning to Rank Dataset and Its Application for Satisfaction-Oriented Ranking},
  author={Li, Yuchen and Zhang, Hao and Zhang, Haojie and Cai, Hengyi and Ma, Xinyu and Wang, Shuaiqiang and Xiong, Haoyi and Ren, Zhaochun and de Rijke, Maarten and Yin, Dawei},
  booktitle=kdd,
  pages={5583--5594},
  year={2025}
}

@inproceedings{li2025rankexpert,
  title={RankExpert: A Mixture of Textual-and-Behavioral Experts for Multi-Objective Learning-to-Rank in Web Search},
  author={Li, Yuchen and Zhang, Hao and Zhang, Yongqi and Cai, Hengyi and Cai, Mingxin and Wang, Shuaiqiang and Xiong, Haoyi and Yin, Dawei and Chen, Lei},
  booktitle=kdd,
  pages={4437--4449},
  year={2025}
}

@article{li2025towards,
  title={Towards AI Search Paradigm},
  author={Li, Yuchen and Cai, Hengyi and Kong, Rui and Chen, Xinran and Chen, Jiamin and Yang, Jun and Zhang, Haojie and Li, Jiayi and Wu, Jiayi and Chen, Yiqun and others},
  journal={arXiv preprint arXiv:2506.17188},
  year={2025}
}

@inproceedings{li2025rankelectra,
  title={RankElectra: Semi-supervised pre-training of learning-to-rank electra for web-scale search},
  author={Li, Yuchen and Xiong, Haoyi and Zhang, Yongqi and Bian, Jiang and Peng, Tianhao and Li, Xuhong and Wang, Shuaiqiang and Kong, Linghe and Yin, Dawei},
  booktitle=kdd,
  pages={2415--2425},
  year={2025}
}

@article{liao2025LearNAT,
  title={LearNAT: Learning NL2SQL with AST-guided Task Decomposition for Large Language Models},
  author={Liao, Weibin and Gao, Xin and Jia, Tianyu and Qiu, Rihong and Zhu, Yifan and Lin, Yang and Chu, Xu and Zhao, Junfeng and Wang, Yasha},
  journal={arXiv preprint arXiv:2504.02327},
  year={2025}
}

@inproceedings{liao2025TPO,
  title={TPO: Aligning Large Language Models with Multi-branch \& Multi-step Preference Trees},
  author={Liao, Weibin and Chu, Xu and Wang, Yasha},
  booktitle=iclr,
  year={2025}
}

@article{liao2025magical,
  title={Magical: Medical Lay Language Generation via Semantic Invariance and Layperson-tailored Adaptation},
  author={Liao, Weibin and Wang, Tianlong and Zhu, Yinghao and Wang, Yasha and Gao, Junyi and Ma, Liantao},
  booktitle=neurips,
  year={2025}
}

@inproceedings{liao2025PAI,
author = {Liao, Weibin and Zhu, Yinghao and Zhang, Zhongji and Wang, Yuhang and Wang, Zixiang and Chu, Xu and Wang, Yasha and Ma, Liantao},
title = {Learnable Prompt as Pseudo-Imputation: Rethinking the Necessity of Traditional EHR Data Imputation in Downstream Clinical Prediction},
year = {2025},
isbn = {9798400712456},
publisher = {Association for Computing Machinery},
address = {New York, NY, USA},
url = {https://doi.org/10.1145/3690624.3709166},
doi = {10.1145/3690624.3709166},
booktitle = kdd,
pages = {765–776},
numpages = {12},
location = {Toronto ON, Canada},
series = {KDD '25}
}

@article{lin2024awq,
  title={Awq: Activation-aware weight quantization for on-device llm compression and acceleration},
  author={Lin, Ji and Tang, Jiaming and Tang, Haotian and Yang, Shang and Chen, Wei-Ming and Wang, Wei-Chen and Xiao, Guangxuan and Dang, Xingyu and Gan, Chuang and Han, Song},
  journal={Proceedings of machine learning and systems (MLSys)},
  volume={6},
  pages={87--100},
  year={2024}
}

@inproceedings{kamath2025pod,
  title={Pod-attention: Unlocking full prefill-decode overlap for faster llm inference},
  author={Kamath, Aditya K and Prabhu, Ramya and Mohan, Jayashree and Peter, Simon and Ramjee, Ramachandran and Panwar, Ashish},
  booktitle={Proceedings of the 30th ACM International Conference on Architectural Support for Programming Languages and Operating Systems (ASPLOS)},
  pages={897--912},
  year={2025}
}

@article{agrawal2023sarathi,
  title={Sarathi: Efficient llm inference by piggybacking decodes with chunked prefills},
  author={Agrawal, Amey and Panwar, Ashish and Mohan, Jayashree and Kwatra, Nipun and Gulavani, Bhargav S and Ramjee, Ramachandran},
  journal={arXiv preprint arXiv:2308.16369},
  year={2023}
}

@inproceedings{zhong2024distserve,
  title={$\{$DistServe$\}$: Disaggregating prefill and decoding for goodput-optimized large language model serving},
  author={Zhong, Yinmin and Liu, Shengyu and Chen, Junda and Hu, Jianbo and Zhu, Yibo and Liu, Xuanzhe and Jin, Xin and Zhang, Hao},
  booktitle={18th USENIX Symposium on Operating Systems Design and Implementation (OSDI)},
  pages={193--210},
  year={2024}
}

@misc{vllm,
  author       = {{Sky Computing Lab, UC Berekely}},
  title        = {{vLLM: Easy, fast, and cheap LLM serving for everyone}},
  year         = {2025},
  howpublished = {\url{https://docs.vllm.ai/en/latest/}},
  note         = {Accessed: 2025-07-31}
}

@misc{triton,
  author       = {{NVIDIA}},
  title        = {{The Triton TensorRT-LLM Backend}},
  year         = {2025},
  howpublished = {\url{https://github.com/triton-inference-server/tensorrtllm_backend}},
  note         = {Accessed: 2025-07-31}
}

@inproceedings{yuan2025native,
  title={Native sparse attention: Hardware-aligned and natively trainable sparse attention},
  author={Yuan, Jingyang and Gao, Huazuo and Dai, Damai and Luo, Junyu and Zhao, Liang and Zhang, Zhengyan and Xie, Zhenda and Wei, YX and Wang, Lean and Xiao, Zhiping and others},
  booktitle=acl,
  year={2025}
}

@article{zheng2024sglang,
  title={Sglang: Efficient execution of structured language model programs},
  author={Zheng, Lianmin and Yin, Liangsheng and Xie, Zhiqiang and Sun, Chuyue Livia and Huang, Jeff and Yu, Cody Hao and Cao, Shiyi and Kozyrakis, Christos and Stoica, Ion and Gonzalez, Joseph E and others},
  journal=neurips,
  volume={37},
  pages={62557--62583},
  year={2024}
}

@article{li2024survey,
  title={A survey on large language model acceleration based on kv cache management},
  author={Li, Haoyang and Li, Yiming and Tian, Anxin and Tang, Tianhao and Xu, Zhanchao and Chen, Xuejia and Hu, Nicole and Dong, Wei and Li, Qing and Chen, Lei},
  journal={arXiv preprint arXiv:2412.19442},
  year={2024}
}

@inproceedings{leviathan2023fast,
  title={Fast inference from transformers via speculative decoding},
  author={Leviathan, Yaniv and Kalman, Matan and Matias, Yossi},
  booktitle=icml,
  pages={19274--19286},
  year={2023},
  organization={PMLR}
}

@inproceedings{ainslie2023gqa,
  title={GQA: Training Generalized Multi-Query Transformer Models from Multi-Head Checkpoints},
  author={Ainslie, Joshua and Lee-Thorp, James and de Jong, Michiel and Zemlyanskiy, Yury and Lebron, Federico and Sanghai, Sumit},
  booktitle=emnlp,
  pages={4895--4901},
  year={2023}
}

@article{dao2022flashattention,
  title={Flashattention: Fast and memory-efficient exact attention with io-awareness},
  author={Dao, Tri and Fu, Dan and Ermon, Stefano and Rudra, Atri and R{\'e}, Christopher},
  journal=neurips,
  volume={35},
  pages={16344--16359},
  year={2022}
}

@article{girija2025optimizing,
  title={Optimizing LLMs for Resource-Constrained Environments: A Survey of Model Compression Techniques},
  author={Girija, Sanjay Surendranath and Kapoor, Shashank and Arora, Lakshit and Pradhan, Dipen and Raj, Aman and Shetgaonkar, Ankit},
  journal={arXiv preprint arXiv:2505.02309},
  year={2025}
}

@inproceedings{wang2024large,
  title={Large search model: Redefining search stack in the era of llms},
  author={Wang, Liang and Yang, Nan and Huang, Xiaolong and Yang, Linjun and Majumder, Rangan and Wei, Furu},
  booktitle=sigir,
  volume={57},
  number={2},
  pages={1--16},
  year={2024},
  organization={ACM New York, NY, USA}
}

@inproceedings{ToolACE,
  title={ToolACE: Winning the Points of LLM Function Calling},
  author={Liu, Weiwen and Huang, Xu and Zeng, Xingshan and Yu, Shuai and Li, Dexun and Wang, Shuai and Gan, Weinan and Liu, Zhengying and Yu, Yuanqing and WANG, Zezhong and others},
  booktitle=iclr,
  year={2025}
}

@inproceedings{Hammer,
  title={Robust function-calling for on-device language model via function masking},
  author={Lin, Qiqiang and Wen, Muning and Peng, Qiuying and Nie, Guanyu and Liao, Junwei and Wang, Jun and Mo, Xiaoyun and Zhou, Jiamu and Cheng, Cheng and Zhao, Yin and others},
  booktitle=iclr,
  year={2025}
}

@misc{bitagent,
  author       = {{BitAgent}},
  title        = {{BitAgent-8B}},
  year         = {2025},
  howpublished = {\url{https://huggingface.co/BitAgent/BitAgent-8B}},
  note         = {Accessed: 2025-05-16}
}

@inproceedings{xlam,
  title={xLAM: A Family of Large Action Models to Empower AI Agent Systems},
  author={Zhang, Jianguo and Lan, Tian and Zhu, Ming and Liu, Zuxin and Hoang, Thai Quoc and Kokane, Shirley and Yao, Weiran and Tan, Juntao and Prabhakar, Akshara and Chen, Haolin and others},
  booktitle=naacl,
  pages={11583--11597},
  year={2025}
}

@article{Qwen,
  title={Qwen technical report},
  author={Bai, Jinze and Bai, Shuai and Chu, Yunfei and Cui, Zeyu and Dang, Kai and Deng, Xiaodong and Fan, Yang and Ge, Wenbin and Han, Yu and Huang, Fei and others},
  journal={arXiv preprint arXiv:2309.16609},
  year={2023}
}

@article{Qwen3,
  title={Qwen3 technical report},
  author={Yang, An and Li, Anfeng and Yang, Baosong and Zhang, Beichen and Hui, Binyuan and Zheng, Bo and Yu, Bowen and Gao, Chang and Huang, Chengen and Lv, Chenxu and others},
  journal={arXiv preprint arXiv:2505.09388},
  year={2025}
}

@inproceedings{BFCL,
  title={The Berkeley Function Calling Leaderboard (BFCL): From Tool Use to Agentic Evaluation of Large Language Models},
  author={Patil, Shishir G and Mao, Huanzhi and Yan, Fanjia and Ji, Charlie Cheng-Jie and Suresh, Vishnu and Stoica, Ion and Gonzalez, Joseph E},
  booktitle=icml,
  year={2025}
}

@article{gpt-4o,
  title={Gpt-4o system card},
  author={Hurst, Aaron and Lerer, Adam and Goucher, Adam P and Perelman, Adam and Ramesh, Aditya and Clark, Aidan and Ostrow, AJ and Welihinda, Akila and Hayes, Alan and Radford, Alec and others},
  journal={arXiv preprint arXiv:2410.21276},
  year={2024}
}

@article{GLM4,
  title={Chatglm: A family of large language models from glm-130b to glm-4 all tools},
  author={GLM, Team and Zeng, Aohan and Xu, Bin and Wang, Bowen and Zhang, Chenhui and Yin, Da and Zhang, Dan and Rojas, Diego and Feng, Guanyu and Zhao, Hanlin and others},
  journal={arXiv preprint arXiv:2406.12793},
  year={2024}
}

@inproceedings{Granite,
  title={Granite-Function Calling Model: Introducing Function Calling Abilities via Multi-task Learning of Granular Tasks},
  author={Abdelaziz, Ibrahim and Basu, Kinjal and Agarwal, Mayank and Kumaravel, Sadhana and Stallone, Matthew and Panda, Rameswar and Rizk, Yara and Bhargav, GP Shrivatsa and Crouse, Maxwell and Gunasekara, Chulaka and others},
  booktitle=emnlp,
  pages={1131--1139},
  year={2024}
}

@article{acharya2025agentic,
  title={Agentic ai: Autonomous intelligence for complex goals--a comprehensive survey},
  author={Acharya, Deepak Bhaskar and Kuppan, Karthigeyan and Divya, B},
  journal={IEEE Access},
  year={2025},
  publisher={IEEE}
}

@inproceedings{wang2025hammerbench,
  title={HammerBench: Fine-Grained Function-Calling Evaluation in Real Mobile Assistant Scenarios},
  author={Wang, Jun and Zhou, Jiamu and Wang, Xihuai and Mo, Xiaoyun and Zhang, Haoyu and Lin, Qiqiang and Jincheng, Jincheng and Wen, Muning and Zhang, Weinan and Peng, Qiuying},
  booktitle=acl_findings,
  pages={3350--3376},
  year={2025}
}

@article{yuan2025mitigating,
  title={Mitigating Tail Latency for on-Device Inference with Load-Balanced Heterogeneous Models},
  author={Yuan, Mu and Zhang, Lan and Duan, Di and Zeng, Liekang and Song, Miao-Hui and Li, Zichong and Xing, Guoliang and Li, Xiang-Yang},
  journal={IEEE Transactions on Mobile Computing},
  year={2025},
  publisher={IEEE}
}

@article{chen2024livemind,
  title={Livemind: Low-latency large language models with simultaneous inference},
  author={Chen, Chuangtao and Zhang, Grace Li and Yin, Xunzhao and Zhuo, Cheng and Schlichtmann, Ulf and Li, Bing},
  journal={arXiv preprint arXiv:2406.14319},
  year={2024}
}

@article{liu2024apigen,
  title={Apigen: Automated pipeline for generating verifiable and diverse function-calling datasets},
  author={Liu, Zuxin and Hoang, Thai and Zhang, Jianguo and Zhu, Ming and Lan, Tian and Tan, Juntao and Yao, Weiran and Liu, Zhiwei and Feng, Yihao and RN, Rithesh and others},
  journal=neurips,
  volume={37},
  pages={54463--54482},
  year={2024}
}

@inproceedings{ning2023all,
  title={All in tokens: Unifying output space of visual tasks via soft token},
  author={Ning, Jia and Li, Chen and Zhang, Zheng and Wang, Chunyu and Geng, Zigang and Dai, Qi and He, Kun and Hu, Han},
  booktitle=iccv,
  pages={19900--19910},
  year={2023}
}

@inproceedings{kong2022spvit,
  title={Spvit: Enabling faster vision transformers via latency-aware soft token pruning},
  author={Kong, Zhenglun and Dong, Peiyan and Ma, Xiaolong and Meng, Xin and Niu, Wei and Sun, Mengshu and Shen, Xuan and Yuan, Geng and Ren, Bin and Tang, Hao and others},
  booktitle=eccv,
  pages={620--640},
  year={2022},
  organization={Springer}
}

@inproceedings{hadsell2006dimensionality,
  title={Dimensionality reduction by learning an invariant mapping},
  author={Hadsell, Raia and Chopra, Sumit and LeCun, Yann},
  booktitle=cvpr,
  volume={2},
  pages={1735--1742},
  year={2006},
  organization={IEEE}
}

@article{oord2018representation,
  title={Representation learning with contrastive predictive coding},
  author={Oord, Aaron van den and Li, Yazhe and Vinyals, Oriol},
  journal={arXiv preprint arXiv:1807.03748},
  year={2018}
}

@inproceedings{he2020momentum,
  title={Momentum contrast for unsupervised visual representation learning},
  author={He, Kaiming and Fan, Haoqi and Wu, Yuxin and Xie, Saining and Girshick, Ross},
  booktitle=cvpr,
  pages={9729--9738},
  year={2020}
}

@inproceedings{liu2022p,
  title={P-Tuning: Prompt Tuning Can Be Comparable to Fine-tuning Across Scales and Tasks},
  author={Liu, Xiao and Ji, Kaixuan and Fu, Yicheng and Tam, Weng and Du, Zhengxiao and Yang, Zhilin and Tang, Jie},
  booktitle=acl,
  pages={61--68},
  year={2022}
}

@inproceedings{li2021prefix,
  title={Prefix-Tuning: Optimizing Continuous Prompts for Generation},
  author={Li, Xiang Lisa and Liang, Percy},
  booktitle=acl_ijcnlp,
  pages={4582--4597},
  year={2021}
}

@inproceedings{minigpt,
  title={MiniGPT-4: Enhancing Vision-Language Understanding with Advanced Large Language Models},
  author={Zhu, Deyao and Chen, Jun and Shen, Xiaoqian and Li, Xiang and Elhoseiny, Mohamed},
  booktitle=iclr,
  year={2024}
}

@article{llava,
  title={Visual instruction tuning},
  author={Liu, Haotian and Li, Chunyuan and Wu, Qingyang and Lee, Yong Jae},
  journal=neurips,
  volume={36},
  pages={34892--34916},
  year={2023}
}

@article{wolf2019huggingface,
  title={Huggingface's transformers: State-of-the-art natural language processing},
  author={Wolf, T},
  journal={arXiv preprint arXiv:1910.03771},
  year={2019}
}

@inproceedings{luohekeep,
  title={Keep the Cost Down: A Review on Methods to Optimize LLM’s KV-Cache Consumption},
  author={Luohe, Shi and Zhang, Hongyi and Yao, Yao and Li, Zuchao and others},
  booktitle=colm,
  year={2024}
}

@inproceedings{qintoolllm,
  title={ToolLLM: Facilitating Large Language Models to Master 16000+ Real-world APIs},
  author={Qin, Yujia and Liang, Shihao and Ye, Yining and Zhu, Kunlun and Yan, Lan and Lu, Yaxi and Lin, Yankai and Cong, Xin and Tang, Xiangru and Qian, Bill and others},
  booktitle=iclr,
  year={2024}
}

@article{mialon2023augmented,
  title={Augmented Language Models: a Survey},
  author={Mialon, Gr{\'e}goire and Dess{\`\i}, Roberto and Lomeli, Maria and Nalmpantis, Christoforos and Pasunuru, Ram and Raileanu, Roberta and Rozi{\`e}re, Baptiste and Schick, Timo and Dwivedi-Yu, Jane and Celikyilmaz, Asli and others},
  journal=tmlr,
  volume={2023},
  year={2023}
}

@article{hsieh2023tool,
  title={Tool documentation enables zero-shot tool-usage with large language models},
  author={Hsieh, Cheng-Yu and Chen, Si-An and Li, Chun-Liang and Fujii, Yasuhisa and Ratner, Alexander and Lee, Chen-Yu and Krishna, Ranjay and Pfister, Tomas},
  journal={arXiv preprint arXiv:2308.00675},
  year={2023}
}

@inproceedings{ruan2023tptu,
  title={Tptu: Task planning and tool usage of large language model-based ai agents},
  author={Ruan, Jingqing and Chen, Yihong and Zhang, Bin and Xu, Zhiwei and Bao, Tianpeng and Mao, Hangyu and Li, Ziyue and Zeng, Xingyu and Zhao, Rui and others},
  booktitle={NeurIPS 2023 Foundation Models for Decision Making Workshop},
  year={2023}
}

@inproceedings{yao2023react,
  title={React: Synergizing reasoning and acting in language models},
  author={Yao, Shunyu and Zhao, Jeffrey and Yu, Dian and Du, Nan and Shafran, Izhak and Narasimhan, Karthik and Cao, Yuan},
  booktitle=iclr,
  year={2023}
}

@article{schick2023toolformer,
  title={Toolformer: Language models can teach themselves to use tools},
  author={Schick, Timo and Dwivedi-Yu, Jane and Dess{\`\i}, Roberto and Raileanu, Roberta and Lomeli, Maria and Hambro, Eric and Zettlemoyer, Luke and Cancedda, Nicola and Scialom, Thomas},
  journal=neurips,
  volume={36},
  pages={68539--68551},
  year={2023}
}

@article{patil2024gorilla,
  title={Gorilla: Large language model connected with massive apis},
  author={Patil, Shishir G and Zhang, Tianjun and Wang, Xin and Gonzalez, Joseph E},
  journal=neurips,
  volume={37},
  pages={126544--126565},
  year={2024}
}

@article{tang2023toolalpaca,
  title={Toolalpaca: Generalized tool learning for language models with 3000 simulated cases},
  author={Tang, Qiaoyu and Deng, Ziliang and Lin, Hongyu and Han, Xianpei and Liang, Qiao and Cao, Boxi and Sun, Le},
  journal={arXiv preprint arXiv:2306.05301},
  year={2023}
}

@article{qian2025userbench,
  title={UserBench: An Interactive Gym Environment for User-Centric Agents},
  author={Cheng Qian and Zuxin Liu and Akshara Prabhakar and Zhiwei Liu and Jianguo Zhang and Haolin Chen and Heng Ji and Weiran Yao and Shelby Heinecke and Silvio Savarese and Caiming Xiong and Huan Wang},
  journal={arXiv preprint arXiv:2507.22034},
  year={2025}
}

@inproceedings{hu2021lora,
  title={LoRA: Low-Rank Adaptation of Large Language Models},
  author={Edward J. Hu and Yelong Shen and Phillip Wallis and Zeyuan Allen-Zhu and Yuanzhi Li and Shean Wang and Lu Wang and Weizhu Chen},
  booktitle=iclr,
  year={2021}
}

\appendix
\section{Algorithm Description}\label{appendix:algorithm}

\begin{algorithm}[!h]
\caption{Dynamic Templating for Function Call Generation}
\label{alg:dynamic_templating}
\begin{algorithmic}[1]
\Require Fine-tuned model \LMS, Initial context $C$, Selected function signature $f_{sig}$
\Ensure Complete and syntactically correct function call string $S_{out}$

\State $S_{out} \gets \text{""}$ \Comment{\textcolor{gray}{Initialize the output string}}
\State $T_{dyn} \gets \text{ConstructDynamicTemplate}(f_{sig})$ \\ \Comment{\textcolor{gray}{e.g., f(arg=<param></param>...)}}
\State $T_{tokens} \gets \text{Tokenize}(T_{dyn})$
\State $i \gets 0$

\While{$i < \text{length}(T_{tokens})$}
    \State $t_{template} \gets T_{tokens}[i]$
    \If{$t_{template} = \textcolor{blue}{<param>}$} \Comment{\textcolor{gray}{LLM Generation Mode}}
        \Loop
            \State $t_{generated} \gets LM_S.\text{generate}(C, S_{out})$
            \If{$t_{generated} = \textcolor{blue}{</param>}$}
                \State \textbf{break} \Comment{End of parameter value, switch back}
            \EndIf
            \State $S_{out} \gets S_{out} + \text{detokenize}(t_{generated})$
        \EndLoop
    \Else \Comment{\textcolor{gray}{Template Injection Mode}}
        \State $S_{out} \gets S_{out} + \text{detokenize}(t_{template})$
    \EndIf
    \State $i \gets i + 1$
\EndWhile

\State \Return $S_{out}$
\end{algorithmic}
\end{algorithm}

\section{Prompts used in This Work}\label{appendix:prompt}

\begin{tcolorbox}[nobeforeafter, 
    title=Prompt for Function Embedding \& Soft Token Distillation via \LML{} , 
    colframe=gray, 
    colback=white, 
    breakable]

You are an expert in composing functions. You are given a question and a set of possible functions. Based on the question, you will need to make one or more function/tool calls to achieve the purpose.\\
If none of the function can be used, point it out. If the given question lacks the parameters required by the function, also point it out.\\
You should only return the function call in tools call sections.\\
\\
If you decide to invoke any of the function(s), you MUST put it in the format of [func\_name1(params\_name1=params\_value1, params\_name2=params\_value2...), func\_name2(params)]\\
You SHOULD NOT include any other text in the response.\\
Here is a list of functions in JSON format that you can invoke.\\
\{functions\}\\
\\
User Query: \{question\}\\

\end{tcolorbox}

\begin{tcolorbox}[nobeforeafter, 
    title=Prompt for Dynamic Templating, 
    colframe=gray, 
    colback=white, 
    breakable]

<|im\_start|>system\\
\# Tools\\
\\
You may call one or more functions to assist with the user query.\\
\\
You are provided with function signatures within <tools></tools> XML tags:\\
<tools>\\
\{functions\}\\
</tools>\\
\\
For each function call, return a json object with function name and arguments within <tool\_call></tool\_call> XML tags, and your answer for the function parameters must be enclosed within <param></param> tags:\\
<tool\_call>\\
\{\{"name": <function-name>, "arguments": <param><args-json-object></param>\}\}\\
</tool\_call><|im\_end|>\\
<|im\_start|>user\\
\{query\}<|im\_end|>\\
<soft\_token>\\
</soft\_token>\\
<|im\_start|>assistant\\
<think>\\
\\
</think>\\
<tool\_call>\\
\{response\}\\
</tool\_call>\\

\end{tcolorbox}

\section{Details of Baselines}\label{appendix:baselines}

The details of the function calling customized LLMs we use is as follows:

\begin{enumerate}
[leftmargin=*,itemsep=0pt,parsep=0.5em,topsep=0.3em,partopsep=0.3em]
    \item \textbf{HammerToolACE (ICLR'25)}: ToolACE~\cite{ToolACE} is an automatic agent-based pipeline designed to generate accurate, complex, and diverse training data for function calling in LLMs. It employs a self-evolution synthesis mechanism and multi-agent dialogue generation, coupled with dual-layer verification, to construct high-quality tool-learning datasets at scale.
    \item \textbf{Hammer (ICLR'25)}: Hammer~\cite{Hammer} is a family of foundation models tailored for on-device function calling, addressing overfitting issues by leveraging augmented datasets and function masking techniques.
    \item \textbf{xLAM (NAACL'25):} xLAM~\cite{xlam} is a series of Large Action Models specifically designed for AI agent tasks, featuring both dense and mixture-of-experts architectures ranging from 1B to 8$\times$22B parameters.
    \item \textbf{Granite (EMNLP'24):} Granite~\cite{Granite} is an open-source large language model specifically designed to enhance function calling capabilities, enabling it to identify, invoke, and interact with external tools and APIs to accomplish complex tasks. 
\end{enumerate}

\section{Exact Match (EM) Accuracy}\label{appendix:em_acc}

Let \( D = \{(X_i, Y_i)\}_{i=1}^N \) be a test set of \( N \) samples, where \( X_i \) is the input and \( Y_i \) is the ground truth list of labels for the \( i \)-th sample. Let \( \hat{Y}_i \) be the list of labels predicted by the model for input \( X_i \).

The Exact Match score for a single prediction \( \hat{Y}_i \) is determined by an indicator function \( \mathbb{I}(\hat{Y}_i = Y_i) \), which is defined as:

\[
\mathbb{I}(\hat{Y}_i = Y_i) =
\begin{cases}
1 & \text{if } \hat{Y}_i = Y_i \\
0 & \text{if } \hat{Y}_i \neq Y_i
\end{cases}
\]

Here, the condition \( \hat{Y}_i = Y_i \) holds true if and only if the two lists have the same length and their corresponding elements are identical.

The overall Exact Match accuracy over the entire test set is the average of these scores across all \( N \) samples. The formula is given by:

\[
\text{Exact Match} = \frac{1}{N} \sum_{i=1}^{N} \mathbb{I}(\hat{Y}_i = Y_i)
\]

The value of this metric ranges from 0 to 1, where 1 indicates that every predicted list perfectly matches its corresponding ground truth list, and 0 indicates that no prediction is an exact match. [11] This metric is particularly useful when the complete correctness of the entire output is crucial.

\section{Additional Experiments}

\subsection{Sensitivity Analysis of Hyperparameter}

\begin{figure}[!h]
     \subfloat[$\alpha$ in the function retriever.\label{fig:alpha}]{\includegraphics[width=0.48\linewidth]{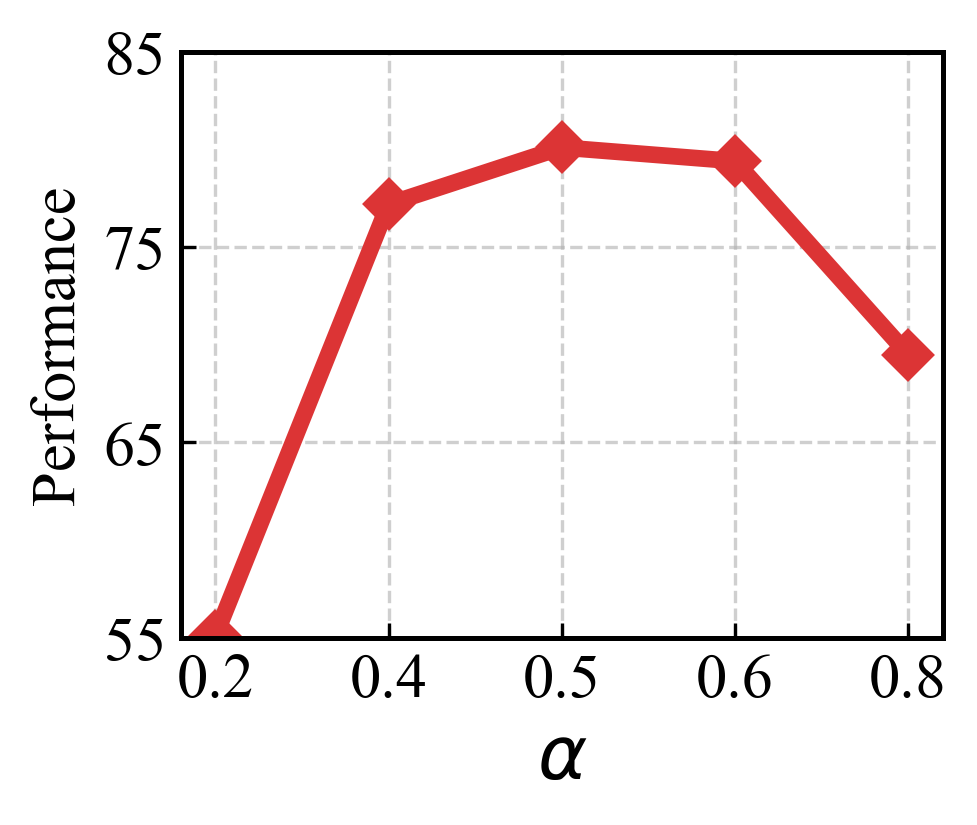}
}\
     \subfloat[Soft token number $k$.\label{fig:k}]{\includegraphics[width=0.48\linewidth]{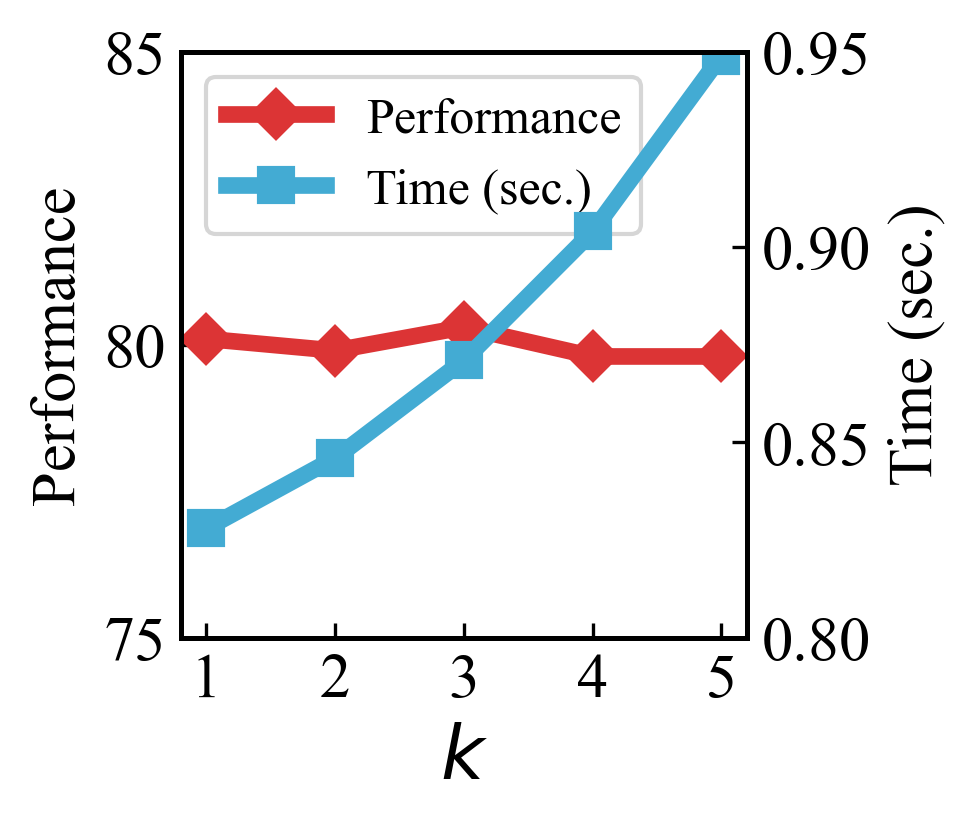}
}
\caption{Sensitivity Analysis.}\label{fig:sensitivity_analysis}
\end{figure}

We provide an analysis of the soft token number $k$ in Fig.~\ref{fig:k}. The experimental results show that the choice of $k$ has minimal impact on model performance; however, since the \LML{} needs to parse more tokens, it leads to higher response latency. This experimental result validates our hypothesis of using the first token to distill user intent.

\subsection{Analysis of Projector Variants}

\begin{table}[!h]
\caption{
Analysis of Various Projector Variants.
}
\label{tab:projector}
\begin{center}
\begin{tabular}{l|cc}
\toprule
\textbf{Projector} & \textbf{Performance} & \textbf{Params} \\
\midrule
MLP (2 layers) & 80.1 & 10.5M \\
LoRA (r=8) & 79.1 & 41K \\
Linear (Ours) & 80.1 & 4.2M \\
\bottomrule
\end{tabular}
\end{center}
\end{table}

We replace the Linear used in Projector with a two-layer MLP and with LoRA~\cite{hu2021lora} (r=8) to analyze the impact of different projector variants on the performance of HyFunc. The experimental results are reported in Table.~\ref{tab:projector}.
The experimental results indicate that our choice of linear achieves a balanced trade-off between parameter count and performance.

\subsection{Analysis of GPU Memory Overhead}

\begin{table}[h]
\caption{
Analysis on GPU Memory Overhead of \TheName{} and baseline models.
}
\label{tab:gpu}
\begin{center}
\begin{tabular}{l|ccc}
\toprule
\textbf{Model} & \textbf{Max.} & \textbf{Time(sec.)} & \textbf{Avg.} \\
\midrule
ToolACE-8B & 16.6G & 1.984 & 16.5G \\
Qwen3-0.6B & 2.3G & 1.006 & 2.3G \\
\midrule
\TheName{} & 18.1G & 0.828 & 4.1G \\
\textit{ - \LML{}} & 16.5G & 0.104 & 16.5G \\
\textit{ - \LMS{}} & 2.3G & 0.724 & 2.3G \\
\bottomrule
\end{tabular}
\end{center}
\end{table}

We provide the GPU Memory of \TheName{} and other baseline models as shown in the Table.~\ref{tab:gpu}. We report the \textit{maximum} (Max.), \textit{average} (Avg.), and \textit{duration} (Time (sec.)) of GPU memory usage during inference. We acknowledge that \TheName{} incurs higher GPU utilization due to the use of a hybrid model; however, this overhead is \textbf{static}. The large model \LML{} is only used for 0.0104 seconds, while most of the inference process is handled by \LMS, resulting in a relatively low average GPU usage for \TheName{}.

\end{document}